\documentclass[conference]{IEEEtran}

\usepackage{cite}
\usepackage{amsmath,amssymb,amsfonts}
\usepackage{algorithmic}
\usepackage{graphicx}
\usepackage{textcomp}
\usepackage{xcolor}
\usepackage{graphicx}
\usepackage{subcaption}
\usepackage{hyperref}
\usepackage{subcaption} 
\usepackage{amsmath}
\usepackage{algorithm}
\usepackage{algorithmic}

\def\BibTeX{{\rm B\kern-.05em{\sc i\kern-.025em b}\kern-.08em
    T\kern-.1667em\lower.7ex\hbox{E}\kern-.125emX}}

\begin{document}

\title{Magnasketch Drone Control}

\author{
    \IEEEauthorblockN{
        Abirami Elangovan\IEEEauthorrefmark{1}, 
        Aatish Gupta\IEEEauthorrefmark{2}, 
        Ashley Kline\IEEEauthorrefmark{3},
    }
    \IEEEauthorblockN{
        Dominique Escandon Valverde\IEEEauthorrefmark{4}, 
        and Scott Wade\IEEEauthorrefmark{5}
    }
    \IEEEauthorblockA{
        Department of Mechanical Engineering, Carnegie Mellon University\\
        Pittsburgh, Pennsylvania, USA\\
        Email: \IEEEauthorrefmark{1}abiramie@andrew.cmu.edu, 
        \IEEEauthorrefmark{2}aatishg@andrew.cmu.edu, 
        \IEEEauthorrefmark{3}ankline@andrew.cmu.edu, \\
        \IEEEauthorrefmark{4}descando@andrew.cmu.edu, 
        \IEEEauthorrefmark{5}scottwad@andrew.cmu.edu
    }
}

\maketitle

\begin{abstract}
The use of Unmanned Aerial Vehicles (UAVs) for aerial tasks and environmental manipulation is increasingly desired. This can be demonstrated via art tasks. This paper presents the development of Magnasketch, capable of translating image inputs into art on a magnetic drawing board via a Bitcraze Crazyflie 2.0 quadrotor. Optimal trajectories were generated using a Model Predictive Control (MPC) formulation newly incorporating magnetic force dynamics. A Z-compliant magnetic drawing apparatus was designed for the quadrotor. Experimental results of the novel controller tested against the existing Position High Level Commander showed comparable performance. Although slightly outperformed in terms of error, with average errors of 3.9 cm, 4.4 cm, and 0.5 cm in x, y, and z respectively, the Magnasketch controller produced smoother drawings with the added benefit of full state control.
\end{abstract}


\section{Introduction}

Unmanned Aerial Vehicles (UAVs) have emerged as versatile tools for various applications, particularly in locations that are difficult to access by humans. Due to the increase of their usage, it is important to improve their ability of aerial manipulation and environmental interaction. One way this has been demonstrated is via art creation. However, utilizing drones as manipulators for drawing or painting presents several unique challenges. Achieving precise and continuous stroke control, maintaining stability during contact, and addressing limitations in drone dynamics require innovative approaches to hardware design, modeling, and control. While existing research has demonstrated creative uses of drones for stippling and calligraphy, these efforts are either computationally and hardware intensive, or lack the ability to generate dynamically feasible trajectories and account for complex manipulator behaviors. 

This project differs from previous work by addressing these limitations. It provides a method of creating optimal trajectories via Model Predictive Control (MPC) for an open-source, accessible drone, the Bitcraze Crazyflie 2.0. Full state control and complex curve following are implemented. A novel magnetic apparatus was designed to allow the drone to create drawings on a magnetic board.

Our contributions included: 
\begin{enumerate}
    \item Modeling of the Crazyflie with simplified magnet-board interaction dynamics
    \item MPC formulation of user input trajectories with contact, control, tracking, and feasibility constraints
    \item Utilization of the Bitcraze High Level and Low Level commanders for successful hardware implementation of differentiable and non-differentiable trajectories
\end{enumerate}

The resulting final controller, which implemented MPC formulation involving magnet dynamics deployed with the Low Level Commander, proved to perform comparably to the existing High Level Commander. Although it was slightly outperformed in terms of error against reference trajectories, it provided smoother, more aesthetically pleasing drawings.

\section{Related Literature}

Previous research has utilized drones for art tasks. In Stippling with Quadrotors \cite{Galea2016} , Galea et al used a Crazyflie with a sponge brush to produce an image by stippling ink on a canvas. This can be seen \href{https://www.youtube.com/watch?v=0irnQVypO7k}{here}, at timestamp 0:20. A motion capture system with a global position Kalman filter was used as their observer. PID controllers were implemented for thrust and hover. A drawback was the lack of orientation control and the ability to draw continuous strokes.

\begin{figure}[hbt!]
    \centering
    \includegraphics[width=0.8\linewidth]{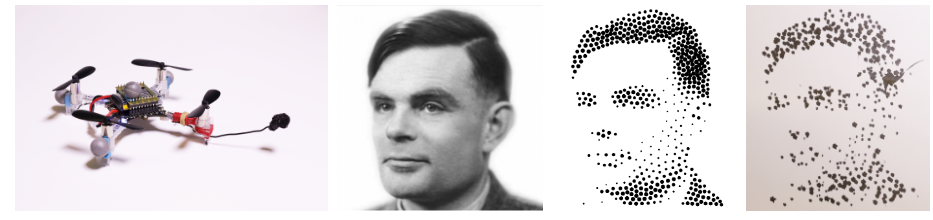}
    \caption{Stippling CrazyFlie}
    \label{fig:stippling}
\end{figure}

In Flying Calligrapher \cite{guo2024flying}, Guo and He et al. produced a hexacopter drone that used a force-sensing sponge brush to produce continuous strokes with ink to create calligraphy. Since the width of the strokes depended on the force applied, the contact-aware trajectory generation involved in this work involved solving non-linear optimization problems. They used a hybrid PID-Impedance controller, with the PID controller controlling motion and the Impedance controller controlling applied force. The limitations of this work are the heavy-duty drone used and the computationally expensive trajectory optimizer.

\begin{figure}[hbt!]
    \centering
    \includegraphics[width=0.5\linewidth]{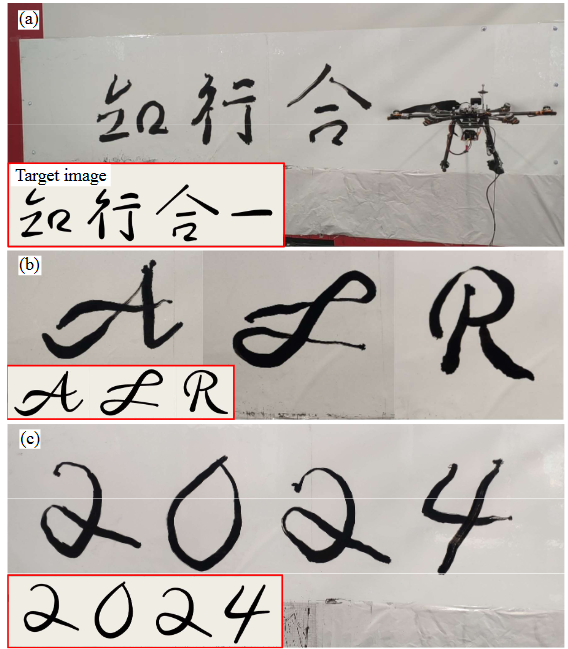}
    \caption{Flying Calligrapher}
    \label{fig:calligrapher}
\end{figure}

Other Notable Works:
\begin{itemize}
    \item \href{https://www.bitcraze.io/2019/03/lighthouse-painting/}{A demo} in which Bitcraze uses an LED to create a long exposure image. \cite{bitcraze_lighthouse_painting}
    \item In \href{https://dl.acm.org/doi/pdf/10.1145/3306306.3328000}{this paper}, Uryasheva et al. used multiple drones to create graffiti. \cite{10.1145/3306306.3328000}
    \item In \href{https://ieeexplore.ieee.org/abstract/document/8379422}{this paper}, Vempati et al. used nonlinear MPC to control the position of a spray painting drone. \cite{vempati}
\end{itemize}

\section{System Modeling}

\subsection{Crazyflie CAD Modeling}

\begin{figure}\label{OneColumn}
\centering
\includegraphics[width=\columnwidth]{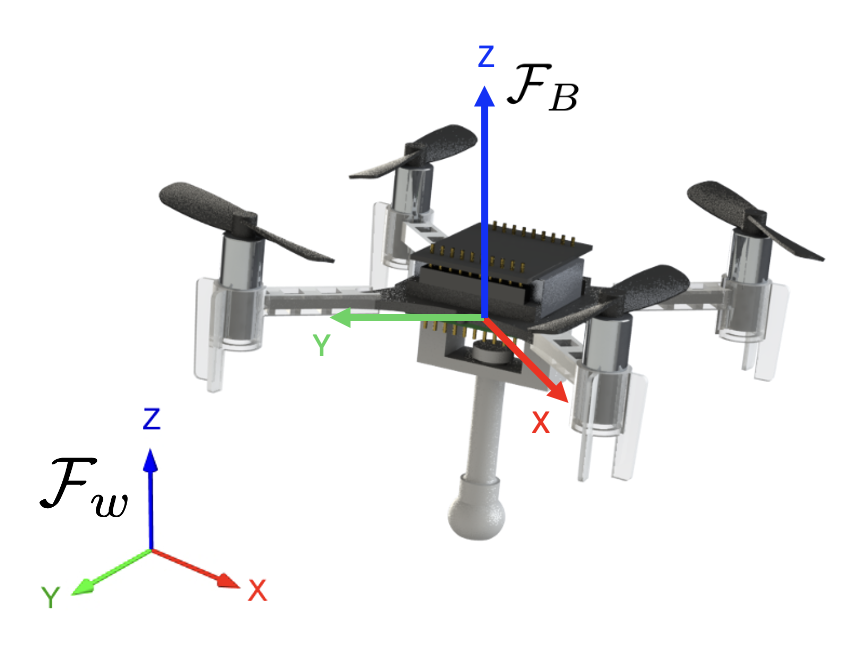}
\caption{The Crazyflie assumes a body-fixed reference frame, $\large \mathcal{F}_B$, at the geometric center of the drone body. A mass-accurate CAD model was generated to provide key model parameters such as mass distribution and the inertia matrix in the body frame. The system resides in the world frame, $\large \mathcal{F}_W$. The CAD rendering was also used in the mesh cat simulation tool.}
\label{fig:fulCAD}
\end{figure}

A precise CAD model of the Bitcraze Crazyflie 2.0 quadrotor was generated to accurately obtain center of mass and inertia matrix data (Fig \ref{fig:fulCAD}). To generate accurate mass distribution, each component of the Crazyflie was measured individually and appropriately modeled. Two frames are defined for this model- the world frame $\large \mathcal{F}_W$ and the body-fixed frame $\large \mathcal{F}_B$. The origin of $\large \mathcal{F}_B$ aligns with the center of mass (and geometric center) of the Crazyflie. To estimate the inertia matrix of the whole system, a few assumptions were made about the magnet payload to simplify the calculation. The magnet payload was modeled as a rigid body that remained fixed in angle below the Crazyflie. The inertia matrix, $\large \mathcal{J}_{rigid body}$, was calculated based off of its center of mass, which was located at 15.383 mm from the top of the rigid body model. The total inertia of the whole system can then be approximated by:

\begin{equation}
    \large \mathcal{J}_{total}= \large \mathcal{J}_{drone} + \large \mathcal{J}_{rigid body}
\end{equation}

Since the origin of the body-fixed frame is positioned at both the geometric center and the center of mass of the drone, there is no need to apply the Parallel Axis Theorem for calculating the moment of inertia. The inertia tensor can be directly calculated from the mass distribution relative to the body frame, as the center of mass coincides with the origin of this frame. It is important to note that because the magnet payload is so light, it hardly changes the mass and inertia matrix, but for accuracy, these updated values were used.

\subsection{Magnet Payload}

Design of the magnetic manipulator was nontrivial and was subject to several competing criteria. Due to the small size and carrying capacity of the Crazyflie platform, the manipulator must be lightweight and minimize the contact force between the drone and the magnetic drawing pad. 

The main method of reducing contact force is to add compliance to the mechanism. Adding compliance heavily reduces certain contact forces while the mechanism remains within the zone of compliance, but the compliance comes at the cost of adding position error to the point of manipulation. Through several iterations, designs were selected primarily to minimize the contact forces, eventually leading to the final design. 

The final design consists of a single 3d printed part which is glued to a battery holder deck. A neodymium magnet is glued into a cavity at the bottom of an arm. The arm is connected to the battery holder deck using a rough ball-and-socket joint, which allows for compliance in z-position, roll, and pitch. This design allows for \~1cm of vertical compliance and \~1cm of radial compliance. The vertical compliance was experimentally verified to be large enough to withstand the z-error in trajectory following using the cascaded PID controller, while the radial compliance was small enough not to distort images too much.

\begin{figure}\label{OneColumn2}
\centering
\includegraphics[width=\columnwidth]{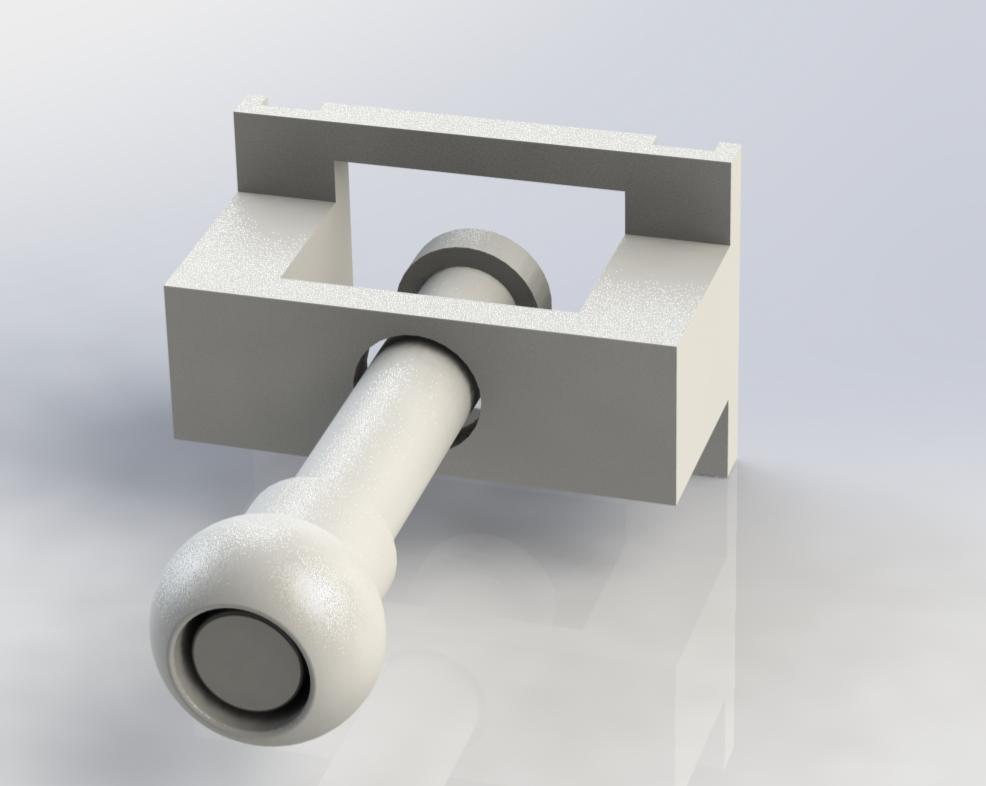}
\caption{The final manipulator design featured a rough ball-and-socket joint to allow for compliance in z-position, roll, and pitch.}
\label{fig:closeupCAD}
\end{figure}

\subsection{Dynamics Equations}\label{AA}
In a standard quadrotor orientation, four propellers rotate
around parallel axes.  We describe the Crazyflie state with 13 states:

\[
{x} = \begin{bmatrix} {r} \\ {q} \\ {v} \\ {\omega} \end{bmatrix}
\]

where $r$ is position, $q$ is orientation in quaternions, $v$ is velocity, and $\omega$ is angular velocity. There are four controls:

\[
{u} = \begin{bmatrix} {u_1} \\ {u_2} \\ {u_3} \\ {u_4} \end{bmatrix}
\]

where $u_i$ is the thrust generated by each $i^th$ propeller.

The Crazyflie system (including the magnet payload) was modeled as a rigid body considering magnet and friction forces, allowing us to use standard Newton-Euler dynamics equations to describe the translational and rotational equations of motion. The translational dynamics (position, velocity, and acceleration) of the drone can be derived from Newton's second law:

\begin{equation}
    \Sigma F = ma
    \label{eq:newton2law}
\end{equation}

where $m$ is the mass of the combined Crazyflie and magnet payload and $F$ are the sum of forces acting on the drone- gravity, thrust force, friction force, and magnet force. Rearranging the terms in Eq. \ref{eq:newton2law} to solve for acceleration, $\dot{v}$, results in:

\begin{equation}
    \dot{v} = \left[ 0, 0, -g \right] + \frac{1}{m} \left( {QF}_{\text{thrust}} + {f}_{\text{magnet}} + {f}_{\text{friction}} \right)
\end{equation}.

The rotation matrix, $Q$, is derived from the quaternion and transforms the body-fixed frame thrust into the world frame. $f_{thrust}$ describes the thrust from each of the propellers and is proportional to $u_i$.

where $k_t$ is the thrust coefficient, which maps the motor input, $u_i$ to the produced thrust. $k_t$ was obtained from Crazyflie and motor specification documents. There are only values in the third row because thrust in the drone body-fixed frame acts only in the z-direction.

The magnet force, $f_{magnet}$, is modeled simply as a downward force acting on the system since the magnet is being pulled toward the magnet board. Based on the maximum pull of the neodymium magnet, which is 0.9lb-f, or approximately 4 Newtons, and the fact that the magnet holder design incorporates a gap between the actual magnet and the board, the downward magnet forces was approximated at:

\begin{equation}
    f_{magnet} = \begin{bmatrix}
        0\\
        0\\
        -2
    \end{bmatrix}
\end{equation}.

The friction force, $f_{friction}$, is modeled simply as a sliding friction:

\begin{equation}
    f_{friction} = \mu m g * \text{sign}(v)
\end{equation}

where $\mu$ is the sliding friction coefficient for ABS plastic, which is what the magnet board is comprised of. 

Strictly speaking, the normal force applied would be the sum of the magnetic force between the magnet and the board, and the difference between the weight of the drone acting downwards and the thrust acting upwards. However, the approximate mass of the Crazyflie is known to be 33g, resulting in a force of 0.3N. This was assumed to be a reasonable approximation of the summation of normal force described. Thus, $mg$ represents the normal force from the drone. 

The friction force opposes velocity and scales with the sliding friction coefficient. A significant assumption was made here that the drone (and magnet payload) is always perpendicular to the magnet board surface. In reality, this is not true, but modeling the friction force this way was more than sufficient to improve overall performance and therefore, this simplification was made. 

Euler's rotation equations can be used to describe the rotational dynamics of the Crazyflie.

\begin{centering}
\begin{equation}
\begin{matrix}
    I_1 \dot{\omega}_1 + (I_3 - I_2) \omega_2 \omega_3 = M_1 \\
I_2 \dot{\omega}_2 + (I_1 - I_3) \omega_3 \omega_1 = M_2 \\
I_3 \dot{\omega}_3 + (I_2 - I_1) \omega_1 \omega_2 = M_3
\end{matrix}
    \label{eq:eulereqn}
\end{equation}
\end{centering}

where $\large \mathcal{J}$ is the combined inertia matrix for the Crazyflie and magnet payload, $\tau$ is the torque from the propellers and $\omega$ cross $\large \mathcal{J}\omega$ is the angular momentum term. Rearranging the terms in Eq. \ref{eq:eulereqn} to solve for angular acceleration, $\dot{\omega}$, results in:

\begin{centering}

\begin{equation}
\begin{matrix}
\dot{\omega}_1 = \frac{M_1 - (I_3 - I_2)\omega_2\omega_3}{I_1}\\
\dot{\omega}_2 = \frac{M_2 - (I_1 - I_3)\omega_3\omega_1}{I_2}\\
\dot{\omega}_3 = \frac{M_3 - (I_2 - I_1)\omega_1\omega_2}{I_3}
\end{matrix}
    \label{eq:eulereqn2}
\end{equation}
\end{centering}
The net torque about each axis is determined by the configuration of the quadrotor and the spinning direction of the rotors. In a standard quadrotor, two rotors spin clockwise and two spin counterclockwise to balance the net angular momentum. Torques about the roll and pitch axes arise from the difference in thrust between motors on opposite sides. Torque about the yaw axis are caused by the drag-induced torques of the rotors. We can write the torques in each axis direction as:

Roll (x-axis): \begin{equation}
                    \tau_x = l(T_3-T_1)
                \end{equation}

Pitch (y-axis): \begin{equation}
                    \tau_y = l(T_4-T_2)
                \end{equation}

Yaw (z-axis): \begin{equation}
                    \tau_z = \begin{bmatrix}
            -l & -l & l & l \\
            -l & l & l & -l \\
            -\dfrac{k_m}{k_t} & \dfrac{k_m}{k_t} & -\dfrac{k_m}{k_t} & \dfrac{k_m}{k_t}
        \end{bmatrix} T
                \end{equation}

where $l$ is the distance from each rotor to the center of mass of the Crazyflie and $T_i$ is the thrust from each propeller:

\begin{equation}
    T_i = k_t * \omega_{i}^2
\end{equation}.

By letting $\omega_{i}^2$ be proportional to the motor input, $u_i$, we can rewrite the torques in each direction as:

\begin{equation}
    f_{thrust} = \begin{bmatrix}
        0 & 0 & 0 & 0\\
        0 & 0 & 0 & 0\\
        k_t & k_t & k_t & k_t
    \end{bmatrix}
    \begin{bmatrix}
        u_1\\
        u_2\\
        u_3\\
        u_4
    \end{bmatrix} = 
    \begin{bmatrix}
        0 \\ 
        0\\
        k_t \Sigma u_i
    \end{bmatrix}
\end{equation}
where $k_m$ is the torque drag coefficient as determined by the motor specifications.  

To write the system in state space, we also need to consider the derivative of the quaternion term in the state. Taking the attitude Jacobian, we can write $\dot{q}$ as:

\begin{equation}
    \dot q = 0.5 * L(q) * H * \omega
\end{equation}

where $L(q)$ maps the angular velocity to the quaternion derivative and $H$ is a "helper" matrix for quaternion math and are described below \cite{jacksonPlanningAttitude2021}.

\begin{equation}
        L = \begin{bmatrix}
        s & -{v}^T \\
        {v} & sI + \text{skew}({v})
        \end{bmatrix}
\end{equation}

where $s$ is the scalar component of the quaternion, $v$ is the vector component of the quaternion, $I$ is the identity matrix, and $skew(v)$ represents the skew symmetric matrix that maps from $\mathbb{R}^{3}$ to $SO3$. 

\begin{equation}
    H = \begin{bmatrix}
    0 & 0 & 0 \\
    1 & 0 & 0 \\
    0 & 1 & 0 \\
    0 & 0 & 1
    \end{bmatrix}
\end{equation}

Putting it all together, we can write the dynamics in state space as:
\small
\begin{equation}
    \begin{bmatrix}
        \dot r\\
        \dot q\\
        \dot v\\
        \dot \omega
    \end{bmatrix} = 
    \begin{bmatrix}
        v\\
        0.5 * L(q) * H * \omega\\
        \begin{bmatrix}
        0\\ 0\\ -g
        \end{bmatrix}
        + \dfrac{Q}{m} 
        \begin{bmatrix}
            0 & 0 & 0 & 0\\
            0 & 0 & 0 & 0 \\
            k_t & k_t & k_t & k_t
        \end{bmatrix} u 
        + f_{magnet} - f_{friction}\\
        J^{-1} -\text{skew}(\omega) J \omega + \begin{bmatrix}
            -l k_t & -l k_t & l k_t & l k_t \\
            -l k_t & l k_t & l k_t & -l k_t \\
            -k_m & k_m & -k_m & k_m
        \end{bmatrix} u
    \end{bmatrix}
\end{equation}
\normalsize
It is important to note that this model does not take into account potential air turbulence or intersecting airflow that could be caused by the tilted propellers. It is assumed to be negligible.

\section{Controller Design}

\subsection{Overview}
The control architecture can be broken into two high-level blocks. The Precompute block involved generating a full state reference trajectory from a series of XYZ points representing a 2D drawing at a fixed Z height. The Onboard block involved utilizing the drone Commander to follow these full state controls. Each of these steps will be expanded on in the following sections. These are summarized below:

\begin{figure}[H]
    \centering
    \includegraphics[width=1\linewidth]{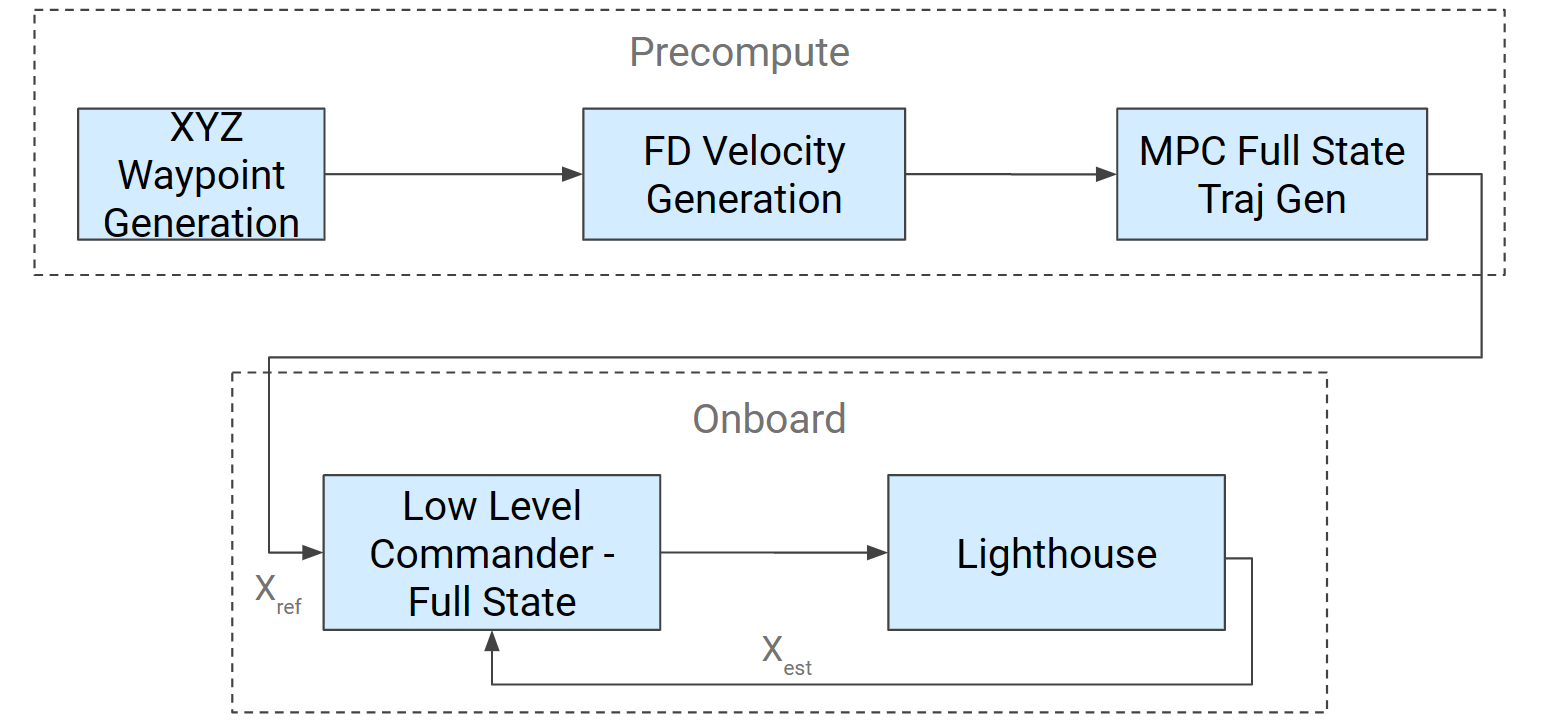}
    \caption{At a high level, control was split between a precompute phase which generated the full state reference trajectory, and an onboard phase, which utilized Bitcraze commanders to follow these setpoints.}
    \label{fig:LL_block_diagram}
\end{figure}

\subsection{Trajectory Generation - Image Based}
Reference trajectory generation required extra processing for this project. Some trajectories, such as the figure-8, were generated using mathematical equations. For image-based trajectories, a standardized process with additional tools were necessary to convert images into waypoints. Coordinator \cite{spotify_coordinator} is a tool that converts SVGs into evenly  spaced points along the outline of the input image and exports these as X, Y coordinates in a CSV file. The points were then normalized to fit the desired dimensions of the drawing board. All SVGs were downloaded from The Noun Project \cite{noun_project}.

Additional points were manually calculated and appended to the trajectory for trajectories requiring the drone to lift off the drawing surface. This is shown in figure \ref{fig:human}.

\subsection{Trajectory Generation - Text Based}

The Coordinator tool proved inadequate for generating text-based trajectories. By creating points along the borders of text, the tool caused each stroke of a letter to be visited twice. To address this issue, computer vision techniques were employed using OpenCV to convert an image of the text into a skeletonized representation. The pixel locations from the skeletonized image were then extracted and transformed into X and Y coordinates. These coordinates were processed using a traveling salesman problem algorithm to generate an optimized reference trajectory, as shown in figure \ref{fig:Trajectory-Generation}.

\begin{figure}
    \centering
    \includegraphics[width=0.75\linewidth]{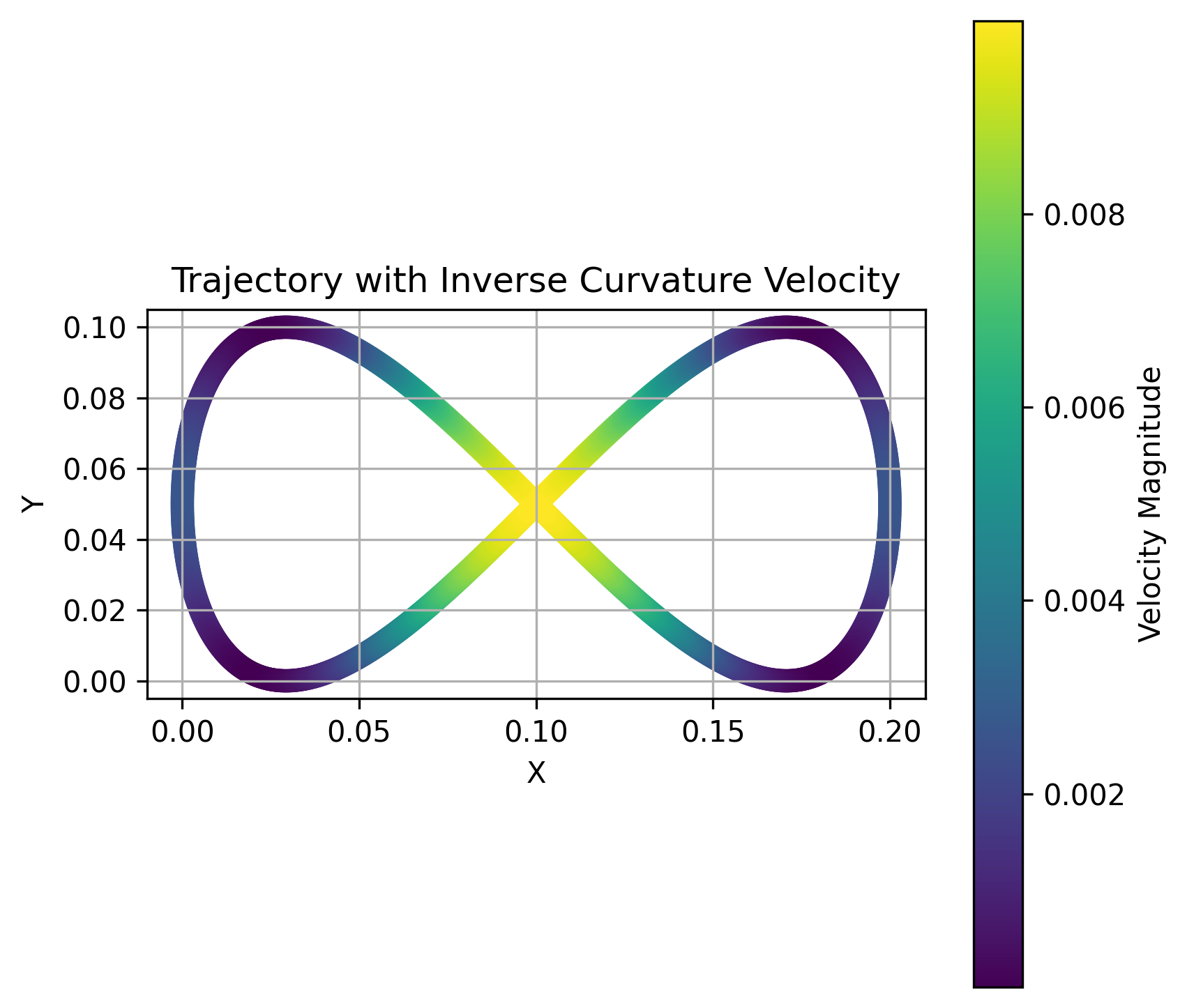}
    \caption{A figure-8 trajectory was the primary trajectory used during initial testing because of its bidirectional nature.}
    \label{fig:f8-curve}
\end{figure}

\begin{figure}
    \centering
    \includegraphics[width=0.75\linewidth]{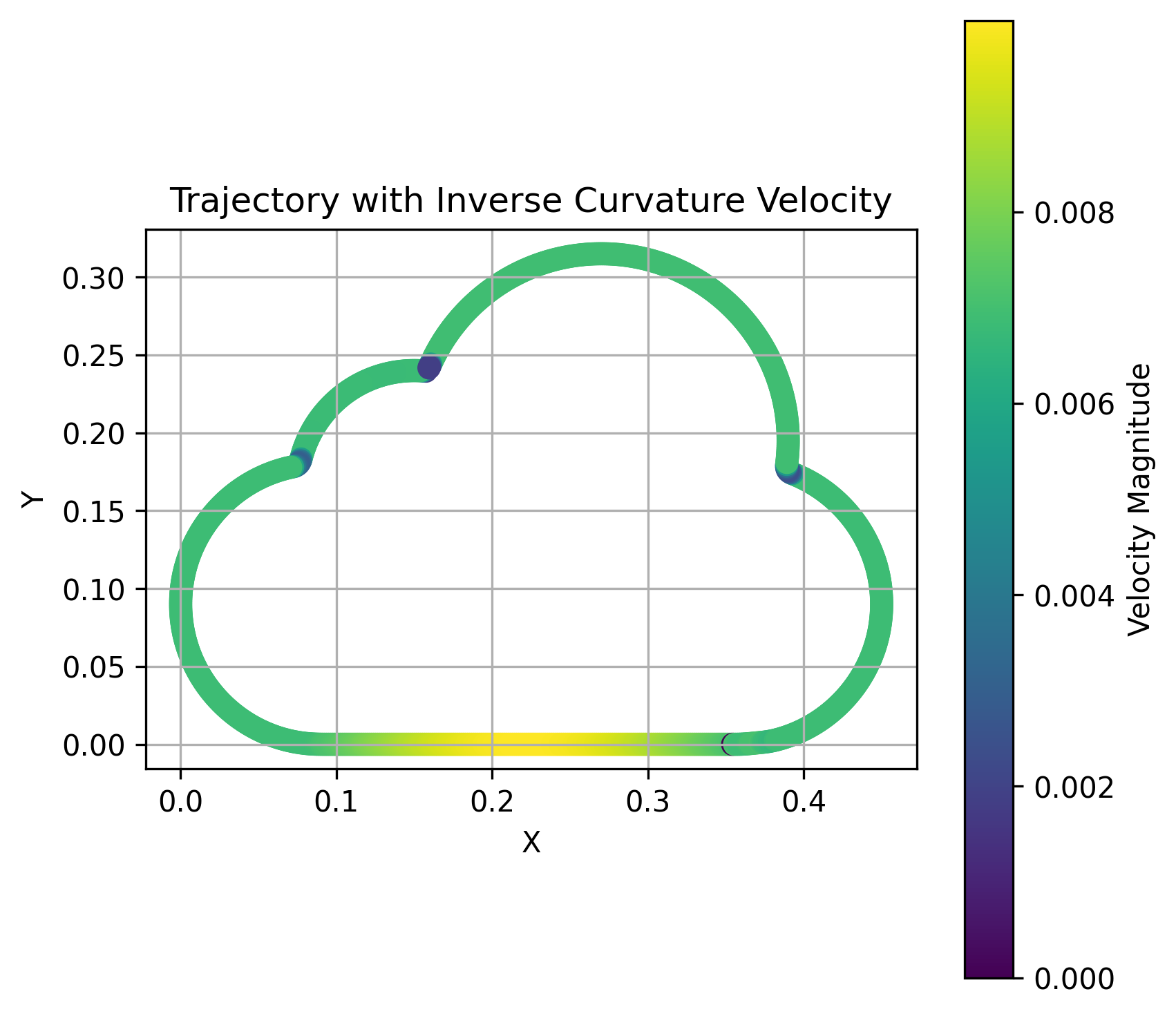}
    \caption{Trajectories calculated using inverse curvature more naturally sped up in straight sections.}
    \label{fig:cloud-FD}
\end{figure}

\begin{figure}
    \centering
    \includegraphics[width=0.75\linewidth]{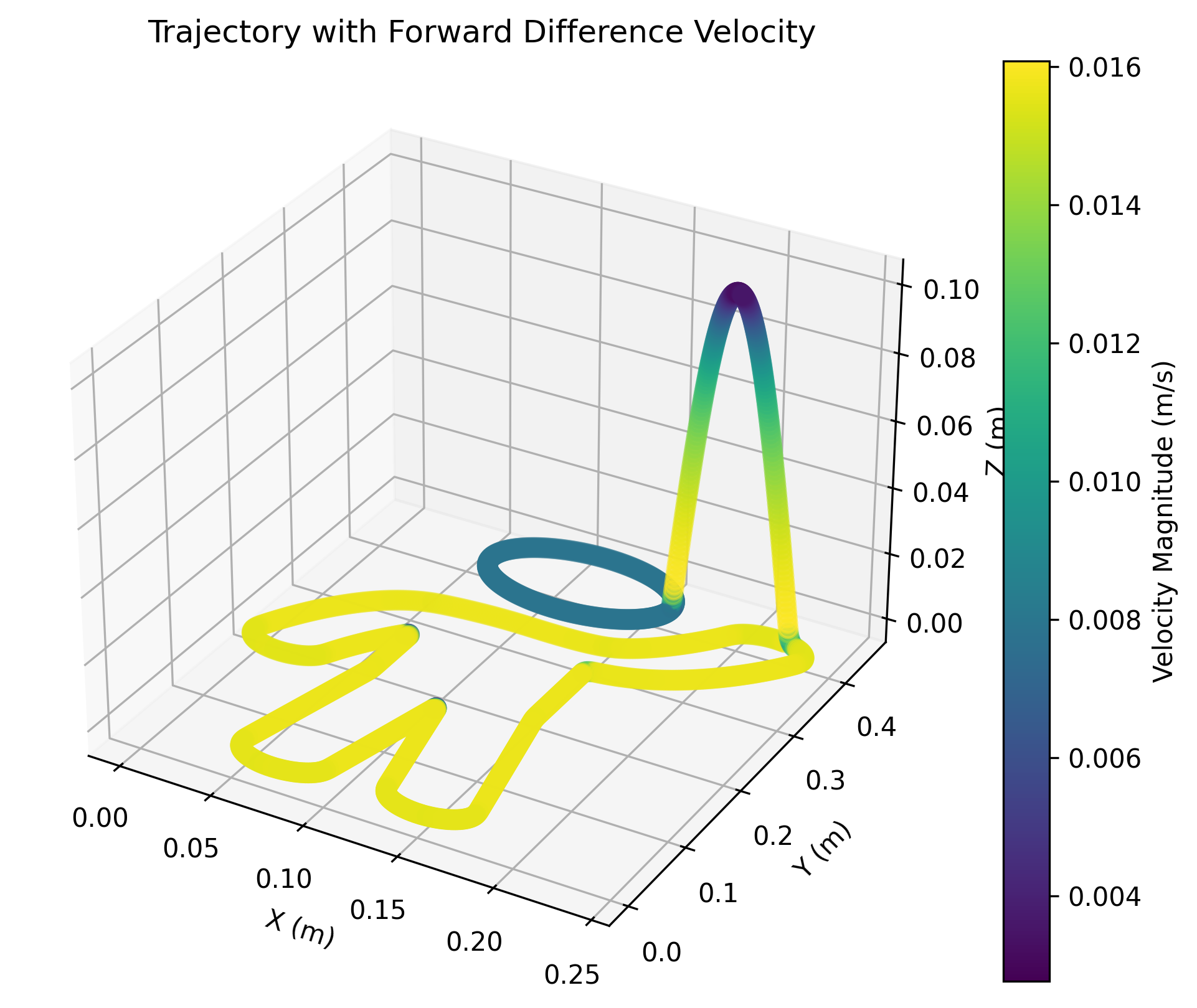}
    \caption{Some trajectories required manual adjustments to promote drone lift off of drawing board for more realistic images.}
    \label{fig:human}
\end{figure}

\subsection{Trajectory Generation - Velocities}
Two methods for generating velocity profiles were tested. Initially, velocities were calculated based on the inverse of curvature at each point along the trajectory. However, since not all images provided continuous curvature data, this method resulted in discontinuous velocities. For smooth trajectories, this was undesirable. To address this, velocities were instead computed using the finite difference method, which produced smoother trajectories. The small distances between points ensured that the velocities remained realistic for the scale of the drawing. For non-smooth (non-differentiable) trajectories such as the cloud, this discontinuity was desirable to slow the drone when a sudden change in direction was required, as shown in figure \ref{fig:cloud-FD}. The inverse curvature method was also desirable in trajectories with long straight lines, because since this method inherently detects straight sections, it was easier to code in the drone "speeding up" on these sections which we found led to more aesthetically pleasing drawings, as shown in figure \ref{fig:f8-curve}

We found that velocities had to be scaled to a maximum velocity of 0.01 m/s in order to get a good solve from the MPC.   

\begin{figure*}[htbp]
    \centering
    \begin{subfigure}[b]{0.32\textwidth}
        \centering
        \includegraphics[width=\linewidth]{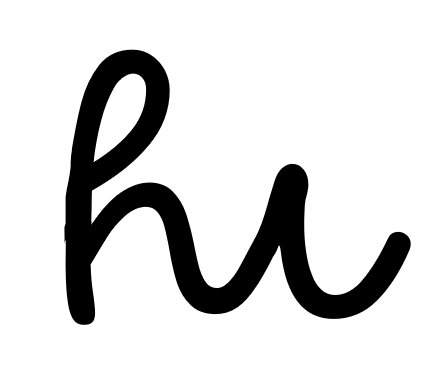}
        \caption{Input image of cursive text, simplified.}
        \label{fig:Input-Text}
    \end{subfigure}
    \begin{subfigure}[b]{0.32\textwidth}
        \centering
        \includegraphics[width=\linewidth]{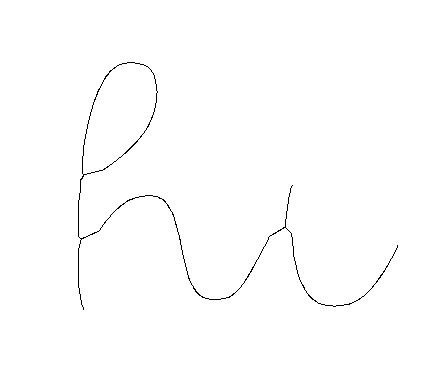}
        \caption{Skeletonized using Zhang Suen thinning.}
        \label{fig:Skeletonized}
    \end{subfigure}
    \begin{subfigure}[b]{0.32\textwidth}
        \centering
        \includegraphics[width=\linewidth]{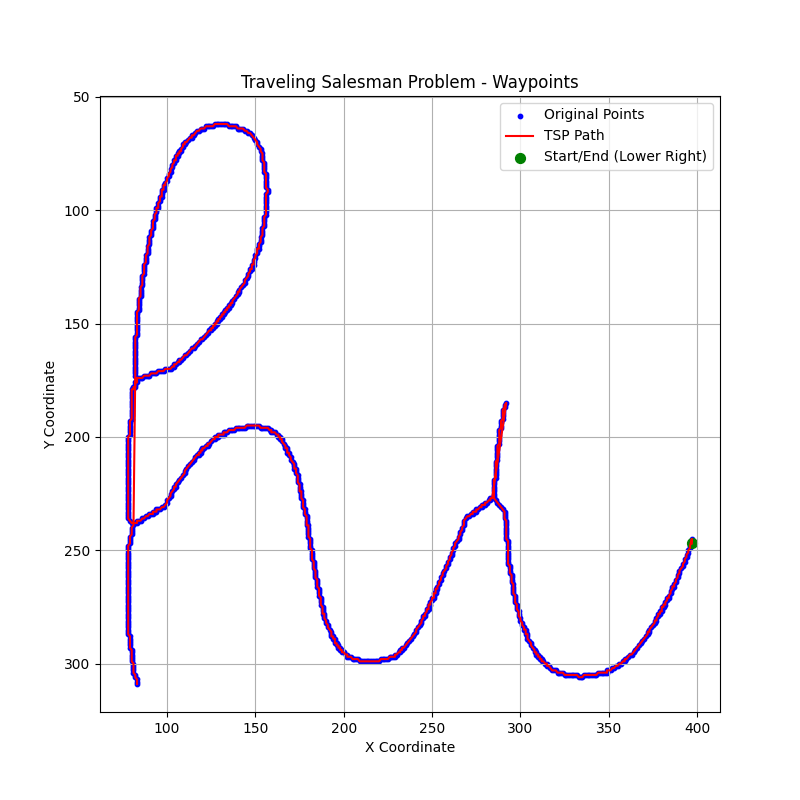}
        \caption{TSP-generated reference trajectory.}
        \label{fig:Traveling-Salesman}
    \end{subfigure}
    \caption{Process for generating a text-based trajectory.}
    \label{fig:Trajectory-Generation}
\end{figure*}

\subsection{Trajectory Optimization}
While the trajectory generation by waypoints was able to provide x, y, z coordinates and approximate velocities to follow, it was critical to establish a trajectory that was dynamically feasible for the drone. In order to establish a dynamically feasible trajectory, convex model predictive control was used. Convex MPC extends the LQR formulation to admit additional convex constraints on the system states and control
inputs, such as motor torque limits. While we could have solved the whole nonlinear trajectory optimization problem for the whole trajectory, we opted to linearize the system around a simple hover state and solve over a horizon. While we were not able to implement an online MPC update for the trajectory on the final control stack, that prospect was a driving factor for selecting the trajectory optimization system we chose for the offline optimized trajectory generation.

\subsubsection{Discretization and Linearization}: In order to use a convex MPC solver (ECOS), the dynamics needed linearized. We did not intend the drone to do any complex aerodynamics and therefore, we performed a single linearization about a hover state equilibrium, ($\bar{X}$, $\bar{U}$). The ForwardDiff package in Julia was used to obtain the Jacobian matrices $A$ and $B$. The equilibrium state vector $[r, q, v, \omega]$ can be described as:

\begin{equation*}
    \bar{X} = [0.0, 0.0, 0.0, 1.0, 0.0, 0.0, 0.0, 0.0, 0.0, 0.0, 0.0, 0.0, 0.0]\mathbf{^T}
\end{equation*}

where the scalar part of the quaternion is listed first in q. The control equilibrium vector was calculated as the amount of thrust for each propeller that was required to counteract gravity and the magnet force of the payload:

\begin{equation*}
    \bar{U}=
        \begin{bmatrix}
        \frac{mg + f_{\text{magnet}}}{4k_t} & \frac{mg + f_{\text{magnet}}}{4k_t} & \frac{mg + f_{\text{magnet}}}{4k_t} & \frac{mg + f_{\text{magnet}}}{4k_t}
        \end{bmatrix}
\end{equation*}.

It was assumed that each propeller thrust for maintaining hover would be equal. 

This nonlinear system was discretized with RK4 then linearized. It is important to note that when using the linearized dynamics, $\Delta$X and $\Delta$U were used in the classic $x_{k+1}= Ax + Bu$, where:
\begin{align*}
    \Delta{X} &= X - \bar{X} \\
    \Delta{U} &= U + \bar{U}
\end{align*}

Any time linearized dynamics were utilized, the $\Delta$ terms were used. $\bar{X}$ and $\bar{U}$ were subtracted and added, respectively, to obtain the true states and controls. To ensure that the discretization and linearization were sound, the norm between the equilibrium state, $\bar{X}$, and the discretized nonlinear dynamics (using the model dt, $\bar{X}$ and $\bar{U}$ was checked. If the equilibrium state satisfies the discretized nonlinear dynamics, the norm will be 0 and this was true for our system. Furthermore, the eigenvalues of the system were checked for stability. To do this, the infinite horizon LQR gain calculated with the linearized dynamics and cost function matrices (described below) was found and the resulting $(A-BK)$ showed all eigenvalues less than 1. 

\subsubsection{MPC Formulation}
Once the system was discretized and linearized, the MPC problem was set up to produce an optimal dynamically feasible trajectory that was given to the onboard controllers on the drone. The optimization problem was set up with the following quadratic cost function:

\begin{align*}
\text{minimize} \quad & J(x, u) = \sum_{i=0}^{N-1} \Big( (x_i - x_{\text{desired},i})^T Q (x_i - x_{\text{desired},i}) \\
& \quad + u_i^T R u_i \Big) \\
& \quad + (x_N - x_{\text{desired},N})^T Q_f (x_N - x_{\text{desired},N})
\end{align*}
\quad \textit{subject to:}
    \[
    \mathbf{X}_{k+1} = \mathbf{A}_k \mathbf{X}_k + \mathbf{B}_k \mathbf{U}_k \quad \forall k = 0, 1, \dots, N
    \]
    \[
    \mathbf{u}_{\min} \leq \mathbf{U}_k \leq \mathbf{u}_{\max} \quad \forall k = 0, 1, \dots, N-1
    \]
    \[
    \mathbf{x}_{0} = \mathbf{x}_{ic}
    \]

where $x_{desired}$ represents the waypoint trajectory that we aim to convert into a dynamically feasible version. $Q$ is a positive semi-definite weighting matrix that penalizes deviations of the state 
$x$ from the desired trajectory, $x_{desired}$, and $R$ is a positive semi-definite weighting matrix that penalizes the magnitude of the control inputs, $u$, to encourage efficient control effort. $Q_f$ is the cost matrix for reaching the terminal state of the trajectory. We add the terminal cost term with a larger $Q_f$ than $Q$ to encourage movement towards the goal state. $Q$ and $R$ were tuned by determining the maximum deviation we wanted to allow from the trajectory it was trying to track. We were strict on the position deviation, enforcing a 1 cm maximum deviation. We were less strict on the velocity deviation since we cared less about following the input velocities really tightly, especially if it would lead to an dynamically infeasible solution. The equations for determining $Q$ and $R$ are listed below.

\begin{equation*}
    Q = \text{diag}\left( \frac{1}{\max\_dev\_x^2} \right)
\end{equation*}

\begin{equation*}
    R = \text{diag}\left( \frac{1}{\max\_dev\_u^2} \right)
\end{equation*}

The cost function is subject to the linearized dynamics, initial condition constraints, and control limit constraints (determined from Bitcraze thrust limit documentation).

\begin{figure*}
    \centering
    \includegraphics[width=0.9\textwidth]{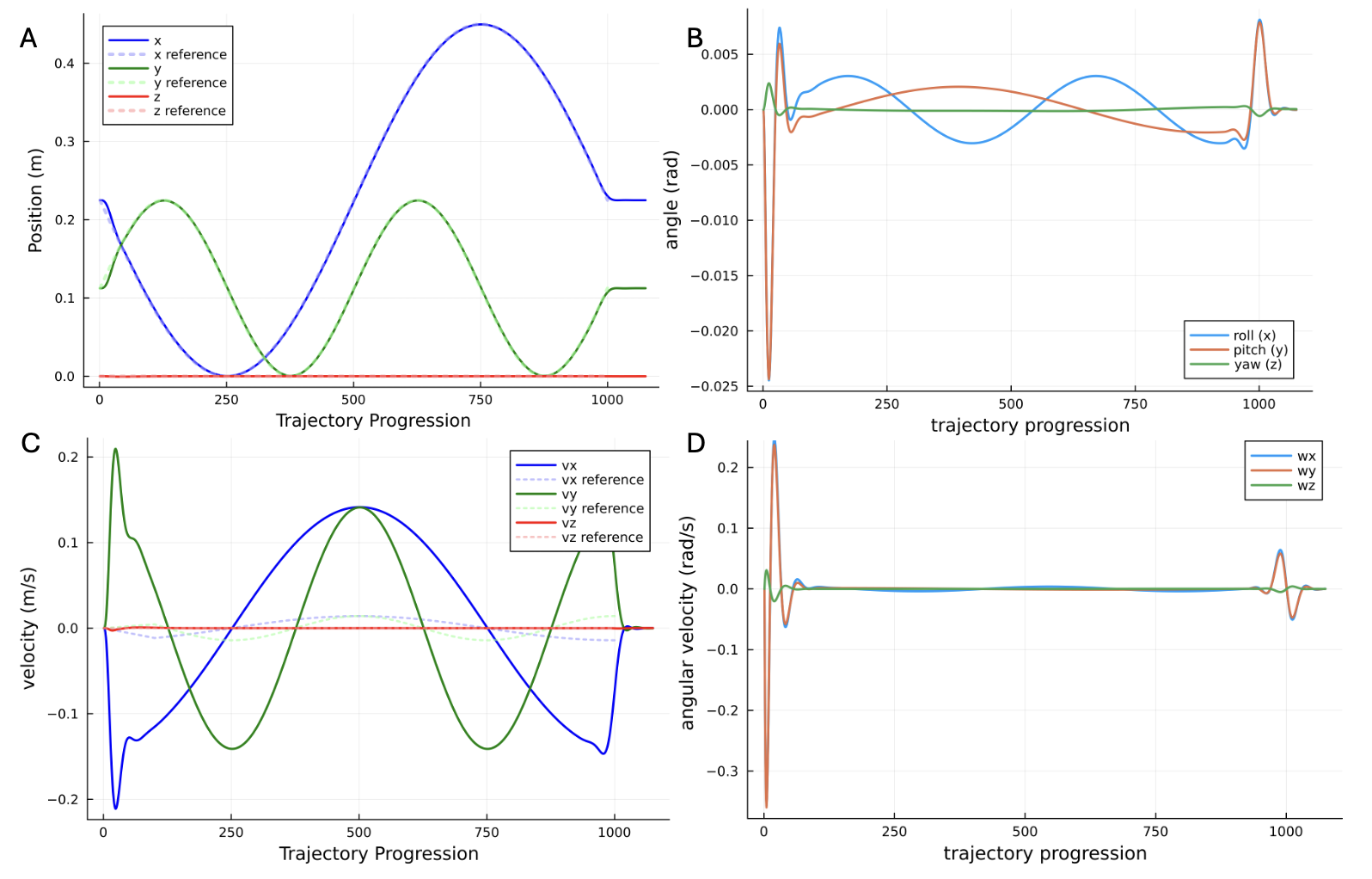}
    \caption{MPC was used to generate an optimal, dynamically feasible trajectory tracked from a drawing of a figure 8 shape. A simulation time step of 0.01 was set, 1000 points were in the drawing trajectory, and an MPC horizon of 75 steps was used. Magnet dynamics considerations were integrated into the nonlinear dynamics. All state curves are smooth due to the differentiable nature of the figure 8 shape. A) Position states track the reference drawing accurately. The MPC was asked to track 0 z-height, as this was set as an offset in low level commander onboard the Crazyflie. B) Orientation states are shown in Euler angles and demonstrate that only small changes in orientation occur to track a 2D drawing in xy. This supports the use of a single hover linearization. C) The velocity states do not track the input drawing as tightly, however the drawing was meant only as an estimate, or warm start. MPC provided dynamically feasible velocity transitions. D) Angular velocity was stable throughout the drawing and did not change except for at the start and end of the drawing when the drone was asked to "instantaneously" start.}
    \label{fig8magnet}
\end{figure*}

\begin{figure*}
    \centering
    \includegraphics[width=0.9\textwidth]{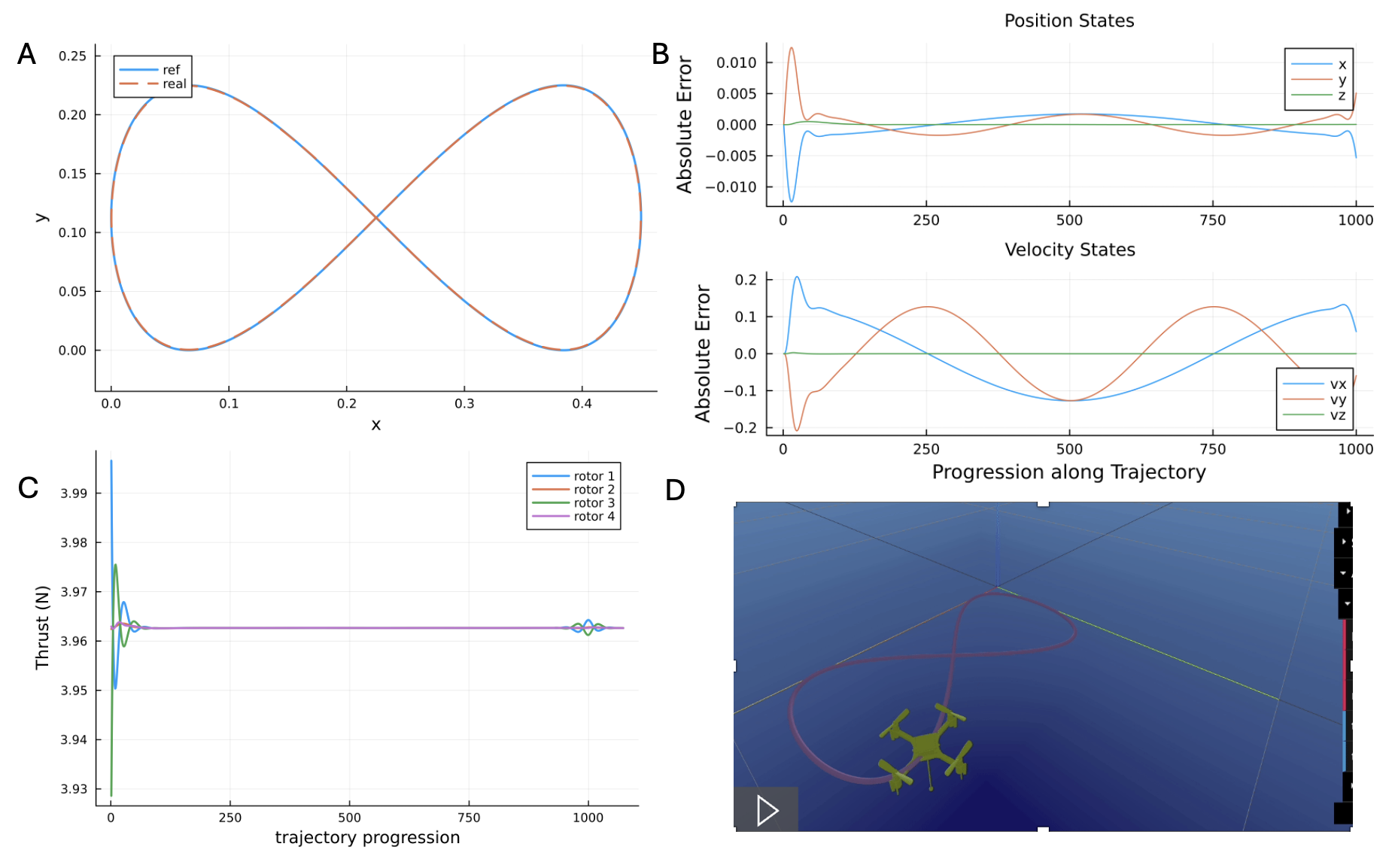}
    \caption{MPC was used to generate an optimal, dynamically feasible trajectory tracked from a drawing of a figure 8 shape. A simulation time step of 0.01 was set, 1000 points were in the drawing trajectory, and an MPC horizon of 75 steps was used. Magnet dynamics considerations were integrated into the nonlinear dynamics. A) An xy state history plot shows accurate position tracking by the MPC trajectory. B) Position and velocity were input as nonzero by the drawing trajectory. The absolute error between the input position and the MPC-generated position states stays below 3mm except at the start of the trajectory when tracking begins. The velocity error between the drawing and the MPC trajectory are higher. Deviations from position and velocity were set in Q as 0.01m and 0.5m/s, respectively. C) Control effort stays around the required thrust to maintain hover. D) A screen shot of the mesh cat simulator shows the drone object tracking the figure 8 drawing.}
    \label{fig8_tracking_image}
\end{figure*}

\begin{figure*}
    \centering
    \includegraphics[width=0.9\textwidth]{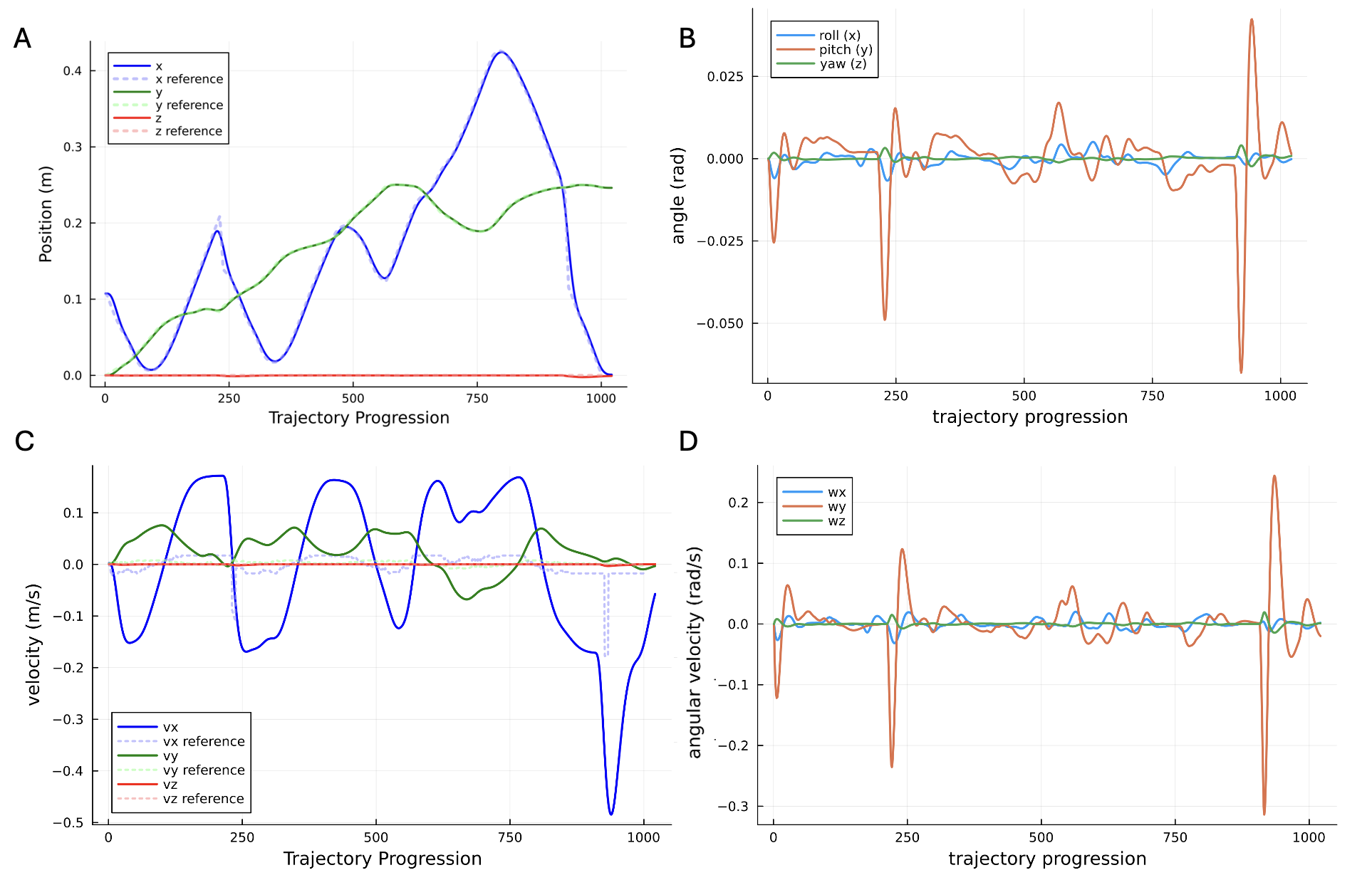}
    \caption{MPC was used to generate an optimal, dynamically feasible trajectory tracked from a drawing of "hi" text. A simulation time step of 0.01 was set, 1001 points were in the drawing trajectory, and an MPC horizon of 20 steps was used. Magnet dynamics considerations were integrated into the nonlinear dynamics. A) Position states track the reference drawing accurately. The MPC was asked to track 0 z-height, as this was set as an offset in low level commander onboard the Crazyflie. B) Orientation states are shown in euler angles and demonstrate that only small changes in orientation occur to track text in xy. This supports the use of a single hover linearization. C) The velocity states do not track the input drawing as tightly, however the drawing was meant only as an estimate, or warm start. MPC provided dynamically feasible velocity transitions. D) Angular velocity variations correlated with changes in orientation. Spikes correlate with sudden direction changes in the non-differentiable text curve.}
    \label{hi_magnet}
\end{figure*}

\begin{figure*}
    \centering
    \includegraphics[width=0.9\textwidth]{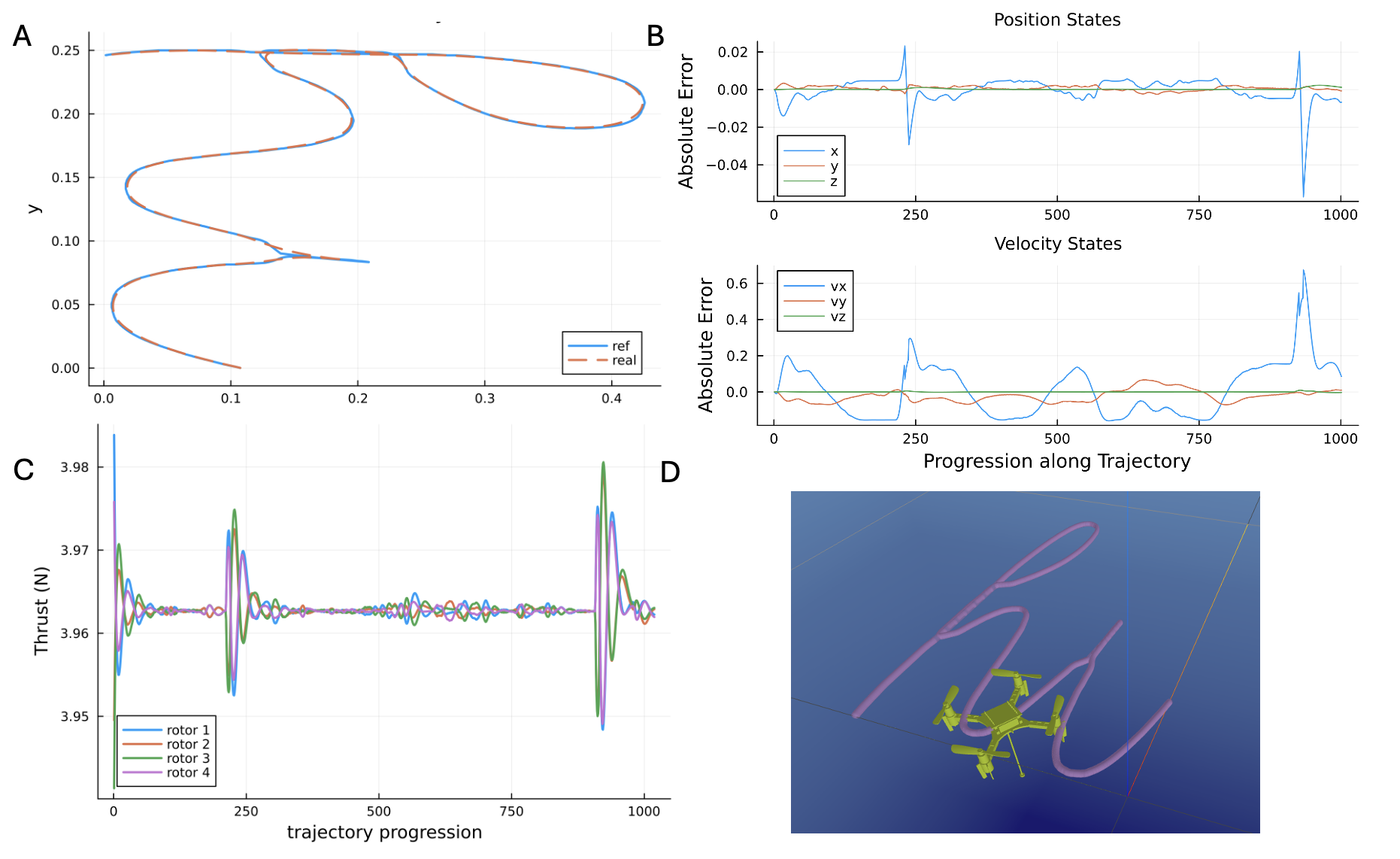}
    \caption{MPC was used to generate an optimal, dynamically feasible trajectory tracked from a drawing of "hi" text. A simulation time step of 0.01 was set, 1001 points were in the drawing trajectory, and an MPC horizon of 20 steps was used. Magnet dynamics considerations were integrated into the nonlinear dynamics. A) An xy state history plot shows accurate position tracking by the MPC trajectory. B) Position and velocity were input as nonzero by the drawing trajectory. The absolute error between the input position and the MPC-generated position states stays below 2mm except at the start of the trajectory when tracking begins. The velocity error between the drawing and the MPC trajectory more variable, as the input velocity was estimating required velocity changes, but MPC found where the dynamics were actually feasible. Deviations from position and velocity were set in Q as 0.01m and 0.5m/s, respectively. C) Control effort maintains around hover thrust, but spikes correlate with sudden direction changes. D) A screen shot of the mesh cat simulator shows the drone object tracking the "hi" text. The MPC trajectory generated the text starting with the "i"}
    \label{hi_tracking_image}
\end{figure*}

The convex MPC solver produced the control vector to solve for the optimal trajectory over the MPC horizon. The first of those controls was taken and passed into the nonlinear dynamics forward rollout to obtain the full state optimal trajectory. The horizon window was then shifted forward by one and the process was repeated. This was done until the entire optimal trajectory to track the waypoint trajectory was generated. We opted to output the full state from the MPC simulation rather than the MPC-generated controls in order to feed the full state directly to the low-level commander of the Crazyflie. Based on our research into the firmware, it was ambiguous as to how the controls were formatted after the PID (i.e. in thrust per motor or thrust and torques) and therefore, we opted to pass along the full state instead to the onboard controllers. 

All trajectory optimization was completed in Julia and a link to the Github is included in the appendix. 

\subsection{Onboard Control and Implementation}
\begin{figure}[H]
    \centering
    \includegraphics[width=1\linewidth]{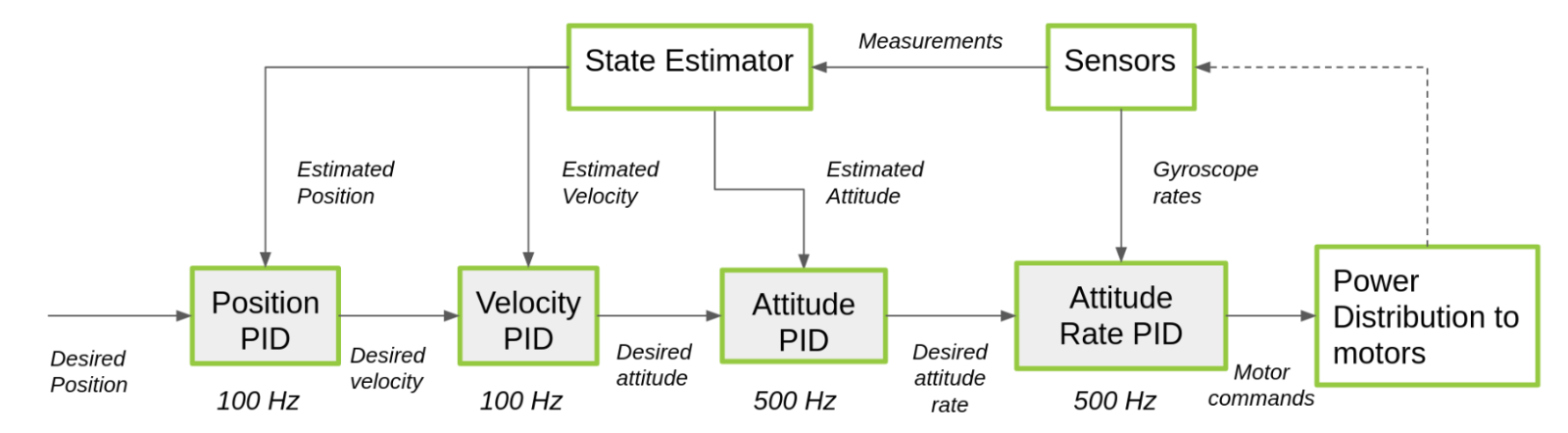}
    \caption{The Bitcraze Cascaded PID was set as the underlying controller for High Level and Low Level Commands. The 100Hz position PID was set as the limit for MPC frequency.}
    \label{fig:bitcraze_PID}
\end{figure}

Once the trajectory optimization was complete, the full state trajectory was stored in a csv. The Onboard controls were then carried out via the Crazyflie Python Library. The Commander framework was utilized, which allows for setpoint control. These all rely on the Bitcraze underlying cascaded PID controller, described in Figure~\ref{fig:bitcraze_PID}. The Lighthouse system provided global position feedback.

1) \textbf{Takeoff:} The trajectory optimization did not include a path for the Crazyflie to take off from the launchpad and reach the first point of the drawing. This was determined to be largely consistent, so only the user input drawing was the trajectory optimization considered. Thus, the \texttt{PositionHLCommander} \texttt{take\_off} commmand was simply used for initial takeoff. Analysis showed that a deadzone existed for the takeoff velocity - below 0.3 m/s would result in instability. 0.5 m/s was determined to be a consistently stable takeoff speed to reach a Z of 0.25m from the launchpad. 

2) \textbf{Establishing Coordinate Frame:} When the Lighthouse is initially calibrated, the {0, 0, 0} is established at the starting position of the drone on top of the launchpad. The trajectory generation, however, assumes a {0, 0, 0} at the ground plane of the drawing. For each trajectory, \texttt{PositionHLCommander} \texttt{go\_to} was used to send an {x, y, z} command that sent the drone to a location centering the drawing on the board. These values were stored as the $x_{offset}$, $y_{offset}$, and $z_{offset}$ values that shifted the coordinate frame from the top of the launchpad to the trajectory generation coordinate frame, simply by summing these values with the trajectory csv {x, y, z} coordinates. All further discussion of trajectory following will assume this coordinate frame shift has been applied. 

3) \textbf{Initial Board Contact:} To reach the first point of the drawing, the \texttt{PositionHLCommander} \texttt{go\_to} command was used to reach the {x, y, z} point. The $z_{offset}$ required significant tuning between trajectories and calibrations to find a value which resulted in consistent contact while preventing the magnet apparatus from hitting its full compliance limit. \texttt{PositionHLCommander} was also discovered to have a non-trivial overshoot and settling time in Z. Thus, to reach the board, \texttt{go\_to} was set to the baseline $z_{offset}$-1.8cm (experimentally found initial overshoot), i.e. 1.8 cm above the board. Control then switched to the \texttt{Low Level Commander}. 

4) \textbf{Stabilizing  Board Contact:} \texttt{Low Level Commander} requires full state commands + acceleration when using the \texttt{send\_full\_state\_setpoint} command. This was wrapped in a \texttt{send\_continuous\_setpoint} function which also took in a duration over which to send the same setpoint command. The settling time was determined to be 1.75 seconds. Thus, to stabilize contact at the board, the first reference state of the trajectory was sent for this time.

5) \textbf{Following Trajectory:} The remainder of the reference trajectory was followed with \texttt{send\_continuous\_setpoint}, with the command duration set to the dt value utilized in MPC. 0.01s, or 100 Hz was found to be sufficient. From Figure~\ref{fig:bitcraze_PID}, this is also the maximum frequency possible for position control. The finite difference method between the current point and the next was used to set acceleration.

6) \textbf{Landing:} Control was switched back to \texttt{PositionHLCommander} to detach from the board and return to the launchpad.

\begin{figure}[H]
    \centering
    \includegraphics[width=1\linewidth]{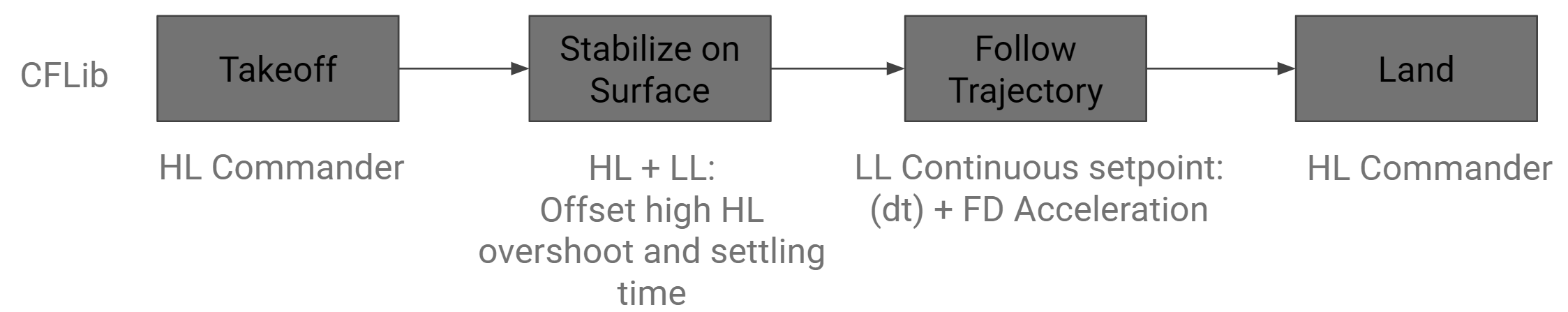}
    \caption{The onboard controller was implemented using cflib. It switches from high level commands for basic takeoff and landing to low level commands for following the full state trajectory. }
    \label{fig:control_block_diagram}
\end{figure}

\section{Simulation}
The first optimal control value was output from MPC and then used in a forward rollout of the nonlinear dynamics to generate the full state associated with using that control. In a simulation loop, the "next state" generated in the forward rollout was used as the initial condition for the next run of convex MPC. The horizon window was shifted forward 1 at every iteration of the simulation loop until the entire trajectory was reached. The reference drawing trajectory that was tracked was padded at the end with the terminal state condition so that MPC could still generate control values when the number of states left in the trajectory was less than the horizon length. Therefore, no variable horizon length was required as the end of the trajectory was reached.  

Figures \ref{fig8magnet}, \ref{fig8_tracking_image}, \ref{hi_magnet}, and \ref{hi_tracking_image} show simulations tracking a figure 8 and "hi" text, respectively. Overall, with only changes in the horizon length, both trajectories, though vastly different, were able to be tracked tightly and the error is low. When tracking error in simulation, we only looked at the position and velocity error since those were the only states that were input as non-zero in the drawing waypoint trajectory. The other states were non-trivial to estimate in the drawing and therefore, we allowed the MPC solver to generate dynamically feasible values for those states. To allow high deviation from the input 0 if necessary, deviation values in Q for orientation and angular velocity were set higher than the allowable deviations for position and velocity. Furthermore, we cared less about deviations in drawing velocity since those were merely estimates for desired velocity at certain parts of each drawing. We did not have exact guesses for when velocities at certain points were no longer feasible and therefore, gave MPC more room to find those. Finally, position deviations were set very strictly to try an limit deviation to under 1 cm. 

It is interesting to note that the simulation is quite generalizable. Only the horizon length for the MPC solve and the number of points in the drawing input were changed. This demonstrates that we are able to accurately simulate drone dynamics and generate feasible behavior to track various types of differentiable and non-differentiable drawings. All trajectories demonstrated in this paper were also tested in simuation without added magnet dynamics to ensure that the added dynamics were not inducing unexpected behavior. The only major difference in simulation between the two was the amount of control required to execute the trajectory. It makes sense that there would be more control effort required to counteract a magnet force and resist sliding friction, but the values produced in sim seemed quite high. However, the hardware could execute the trajectory fairly well irregardless.  

The model parameters used to generate the MPC trajectory are shown in Table \ref{tab:parameters} \cite{alavilliTinyMPCModelPredictiveControl2024}.

\begin{table}[ht] 
\centering
\caption{Model Parameters}
\label{tab:parameters}
\begin{tabular}{|p{1.0cm}|p{3.0cm}|p{2.0cm}|p{1.0cm}|}
\hline
\textbf{Symbol} & \textbf{Description} & \textbf{Value} & \textbf{Units} \\
\hline
$m$ & Mass & 0.033885 & kg \\
\hline
$k_t$ & Thrust coefficient & $1.47 \times 10^{-1}$ &  \\
\hline
$k_m$ & Torque moment coefficient & $1.18 \times 10^{-4}$ &  \\
\hline
$\mathcal{J}$ & Inertia matrix & \parbox{5cm}{$\text{diag}[1.66 \times 10^{-5},$\\$1.66 \times 10^{-5},$\\$2.93 \times 10^{-5}]$} & $\text{kg·m}^2$ \\
\hline
$g$ & Gravity & 9.8 & m/s\(^2\) \\
\hline
$\ell$ & Arm length (along diagonal) & $0.046/\sqrt{2}$ & m \\
\hline
$dt$ & Time step & 100 & Hz \\
\hline
\end{tabular}
\end{table}

A simulator was built in mesh cat to observe tracking accuracy between the waypoint trajectory and the MPC-created trajectory. See the Appendix for supplemental figures showing demonstrations of additional shapes. For another simple, differentiable shape, we tracked a circle. for non-differentiable shapes, we drew a cloud, a human, and a cat. In simulation, even when non-differentiable, the MPC trajectory can accurately track the drawing. 

\section{Hardware Implementation and Testing}

Three methods were compared. 

1) \textbf{XYZ + \texttt{PositionHLCommander}:} This was used as a baseline, with only initial {x, y, z} coordinates generated from coordinator input fed into the commander, with a set velocity of 0.075 m/s.

2) \textbf{MPC + \texttt{LLCommander}:} The full state output from the non-magnet dynamics MPC formulation was fed into low level commander full state setpoint control

3) \textbf{MPC with Magnet Dynamics + \texttt{LLCommander}:} Magnet dynamics were incorporated into MPC formulation and fed into low level commander. 

\begin{figure}[H]
    \centering
    \includegraphics[width=1\linewidth]{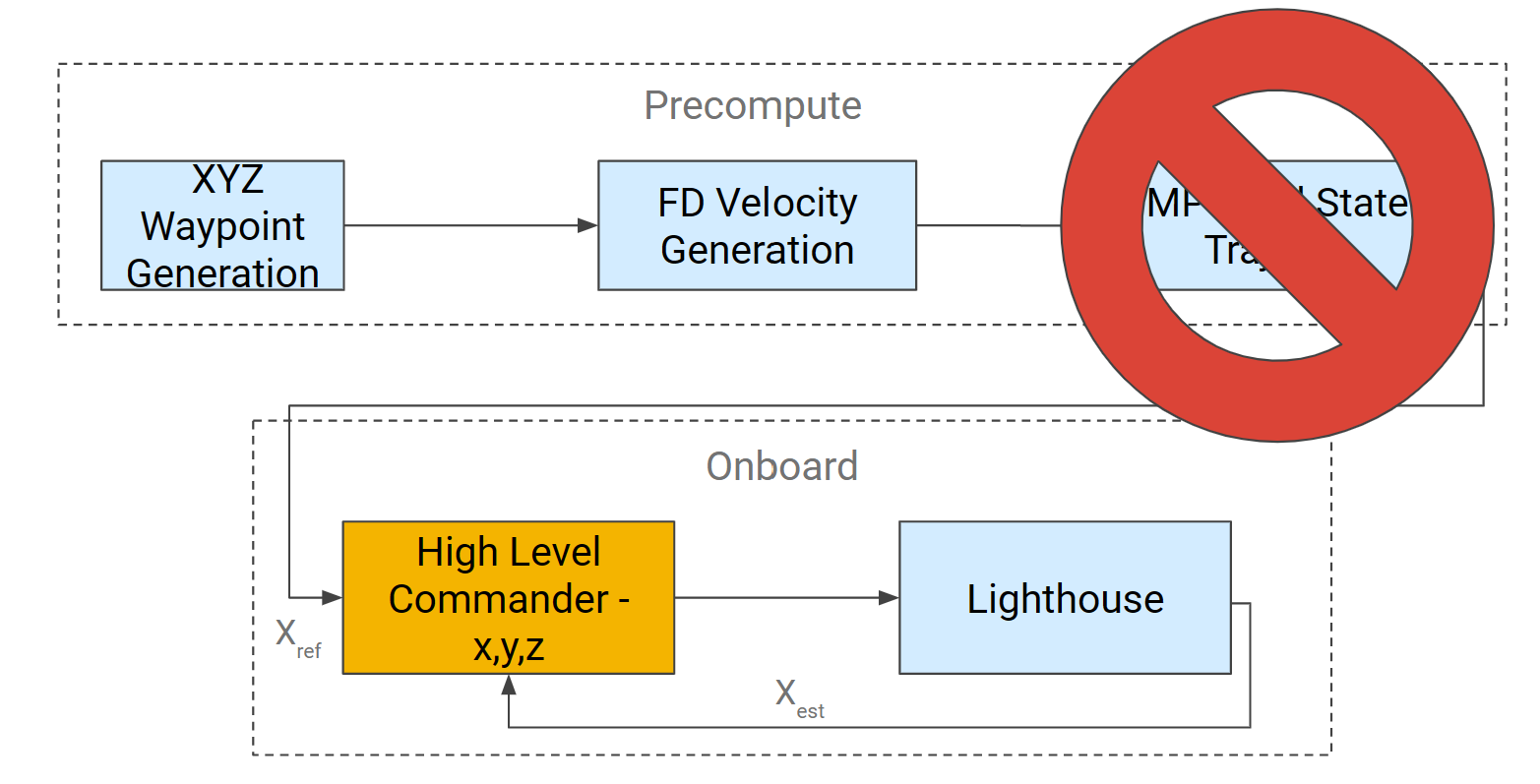}
    \caption{The baseline test removed trajectory optimization and instead only commanded x, y, z through HL Commander.}
    \label{fig:HL_block_diagram}
\end{figure}

The difference between methods 1) and 2) or 3) are evident by comparing Figure~\ref{fig:HL_block_diagram} and Figure~\ref{fig:LL_block_diagram}.

These methods were tested on the figure-8, circle, and cloud trajectories.

\section{Demo Results}

The error across trajectories was analyzed. This was done by taking the average of the absolute values of the error between the reference trajectory and true position at each setpoint. The average and standard deviation across all trajectories is summarized below:

\begin{table}[h!]
    \centering
    \begin{tabular}{|l|c|c|c|}
        \hline
        \textbf{Controller} & \textbf{X $\pm$ SD (cm)} & \textbf{Y $\pm$ SD (cm)} & \textbf{Z $\pm$ SD (cm)} \\ \hline
        HL                   & 2.79 $\pm$ 0.30             & 2.91 $\pm$ 0.22             & 0.46 $\pm$ 0.10             \\ \hline
        No Dynamics          & 4.18 $\pm$ 1.00             & 4.32 $\pm$ 1.36             & 0.54 $\pm$ 0.26             \\ \hline
        Dynamics             & 3.95 $\pm$ 0.66             & 4.43 $\pm$ 1.15             & 0.51 $\pm$ 0.19             \\ \hline
    \end{tabular}
    \caption{Average controller error across trajectories. HL Commander outperforms the LL implementations. However, incorporating magnet dynamics appears to improve LL performance.}
    \label{tab:error_stdev_combined}
\end{table}

Evidently, the baseline High Level Commander implementation results in lower error than the Low Level implementations following the full state optimal reference trajectory. However, including magnet dynamics into the MPC formulation appeared to improve the performance of the LL controller.

A visualization script was developed to observe the trajectory further. This, along with the actual drawings on the board, provided greater insight into the controller performances. The results for the figure-8 in particular will be described, but the findings were similar across trajectories. Figures for the other trajectories can be found in the Appendix. 

\subsection{Figure-8}

\begin{table}[h!]
    \centering
    \begin{tabular}{|l|c|c|c|}
        \hline
        \textbf{Controller} & \textbf{$x_{error}$ (cm)} & \textbf{$y_{error}$ (cm)} & \textbf{$z_{error}$ (cm)} \\ \hline
        HL Commander         & 2.7112                  & 3.0767                  & 0.3595                  \\ \hline
        No Dynamics   & 4.9282                  & 5.8952                  & 0.812                   \\ \hline
        Magnet Dynamics      & 4.2868                  & 5.7235                  & 0.4739                  \\ \hline
    \end{tabular}
    \caption{Controller average error comparison for figure-8, showing the least error from HL Commander, followed by LL Commander with magnet dynamics, and lastly LL Commander without magnet dynamics.}
    \label{tab:fig8_errors}
\end{table}

\begin{figure}[H]
    \centering
    \begin{subfigure}[b]{0.32\linewidth}
        \centering
        \includegraphics[width=\linewidth]{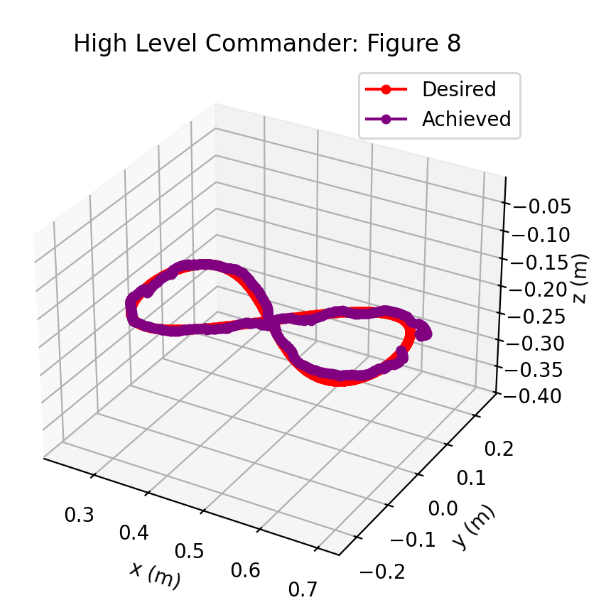}
        \caption{HL Commander}
        \label{fig:fig8_HL}
    \end{subfigure}
    \hfill
    \begin{subfigure}[b]{0.32\linewidth}
        \centering
        \includegraphics[width=\linewidth]{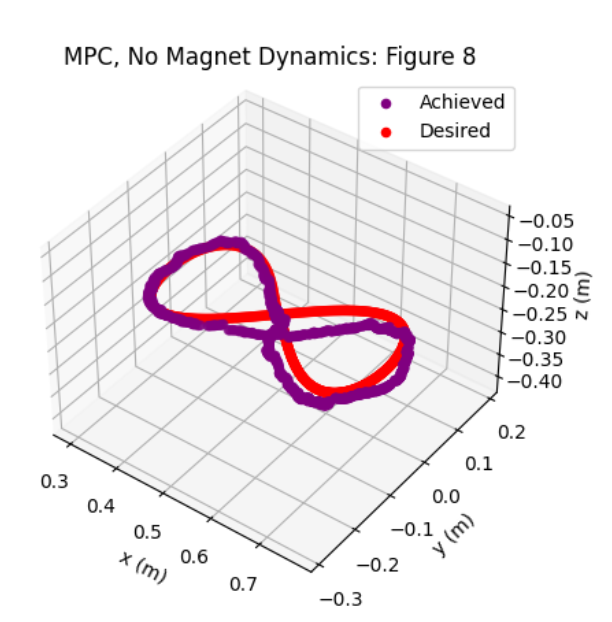}
        \caption{No Dynamics}
        \label{fig:fig8_no_mag}
    \end{subfigure}
    \hfill
    \begin{subfigure}[b]{0.32\linewidth}
        \centering
        \includegraphics[width=\linewidth]{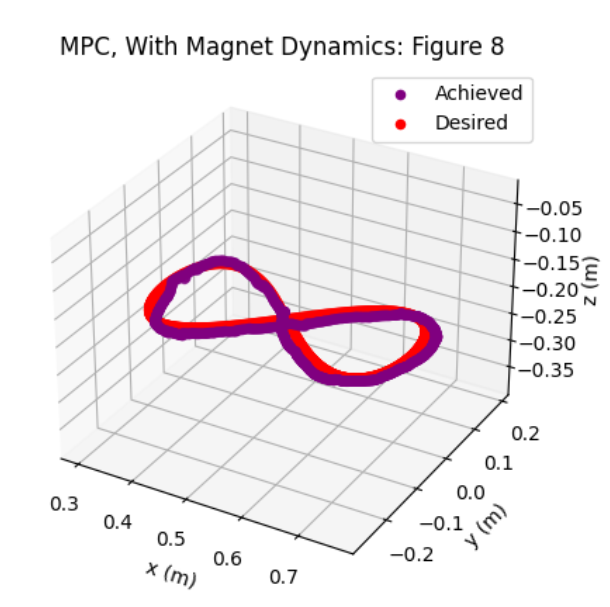}
        \caption{Magnet Dynamics}
        \label{fig:fig8_magnet}
    \end{subfigure}
    \caption{Visualization of the HL Commander, LL Commander without magnet dynamics, and LL Commander with magnet dynamics controllers. These show that although HL Commander results in the least quantified error, incorporating the magnet dynamics actually resulted in smoother drawings.}
    \label{fig:combined_fig8}
\end{figure}

\begin{figure}[H]
    \centering
    \begin{subfigure}[b]{0.32\linewidth}
        \centering
        \includegraphics[width=\linewidth]{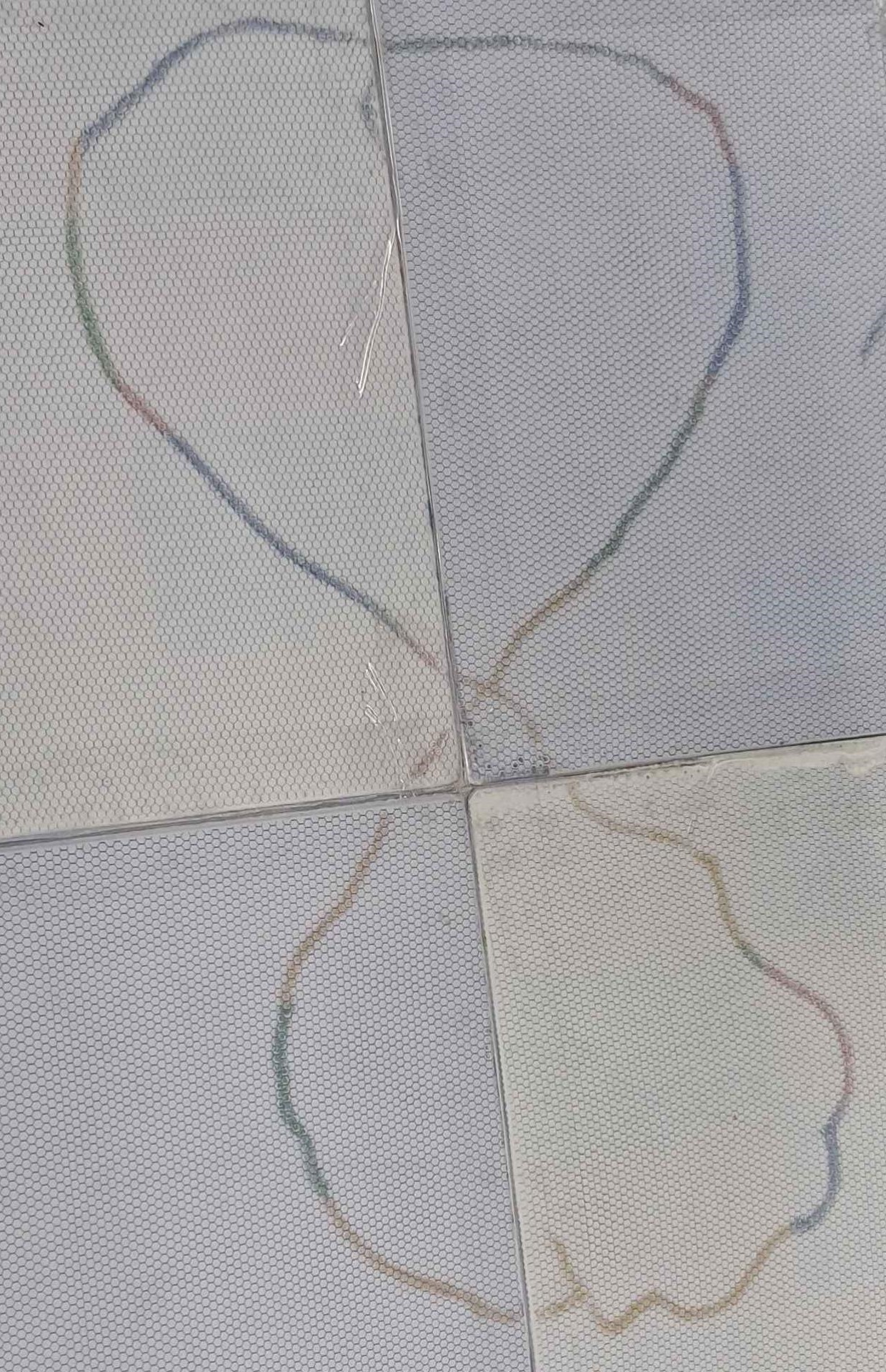}
        \caption{HL Commander}
        \label{fig:gih_HL}
    \end{subfigure}
    \hfill
    \begin{subfigure}[b]{0.32\linewidth}
        \centering
        \includegraphics[width=\linewidth]{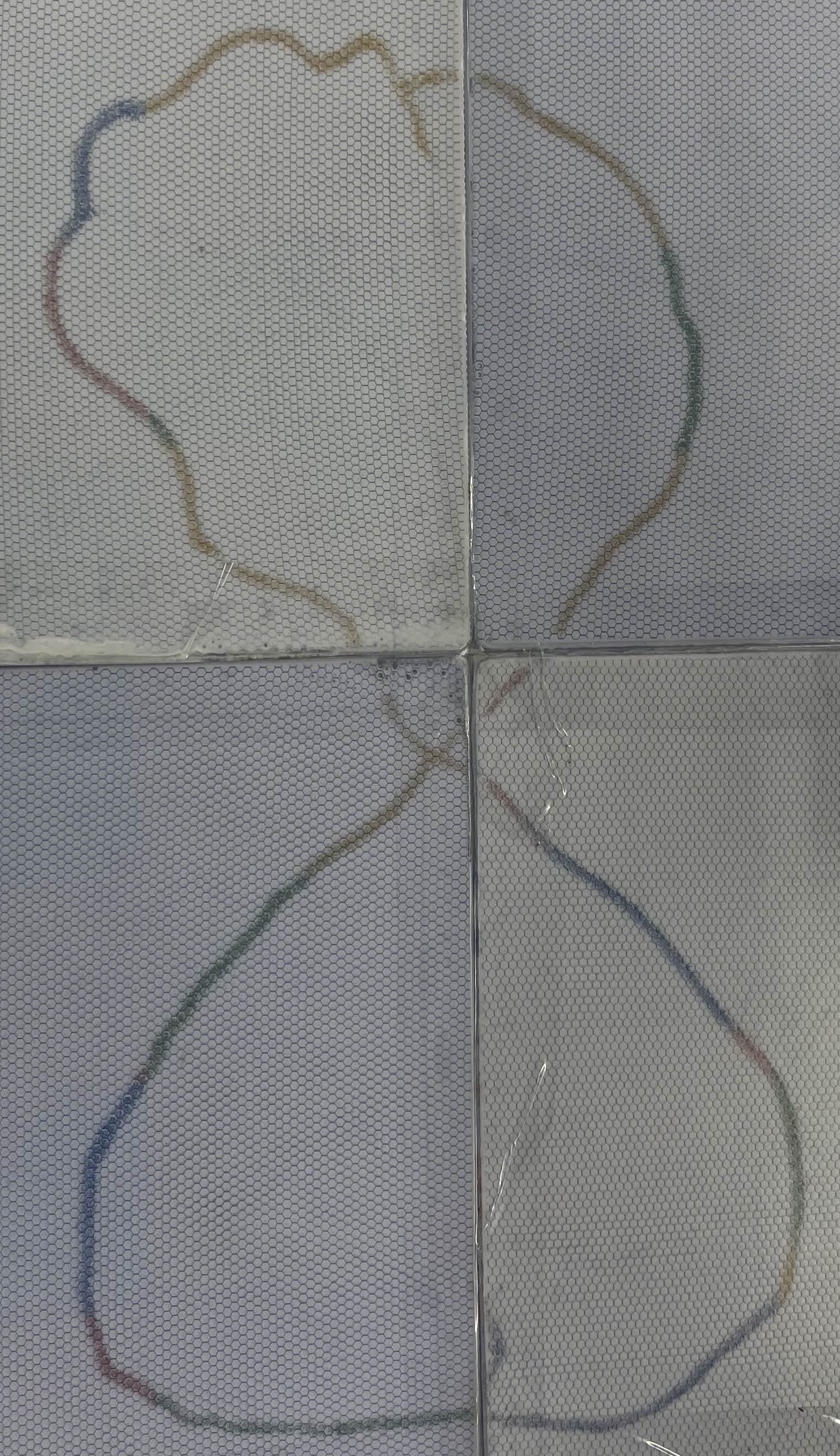}
        \caption{No Dynamics}
        \label{fig:fiasda_no_mag}
    \end{subfigure}
    \hfill
    \begin{subfigure}[b]{0.32\linewidth}
        \centering
        \includegraphics[width=\linewidth]{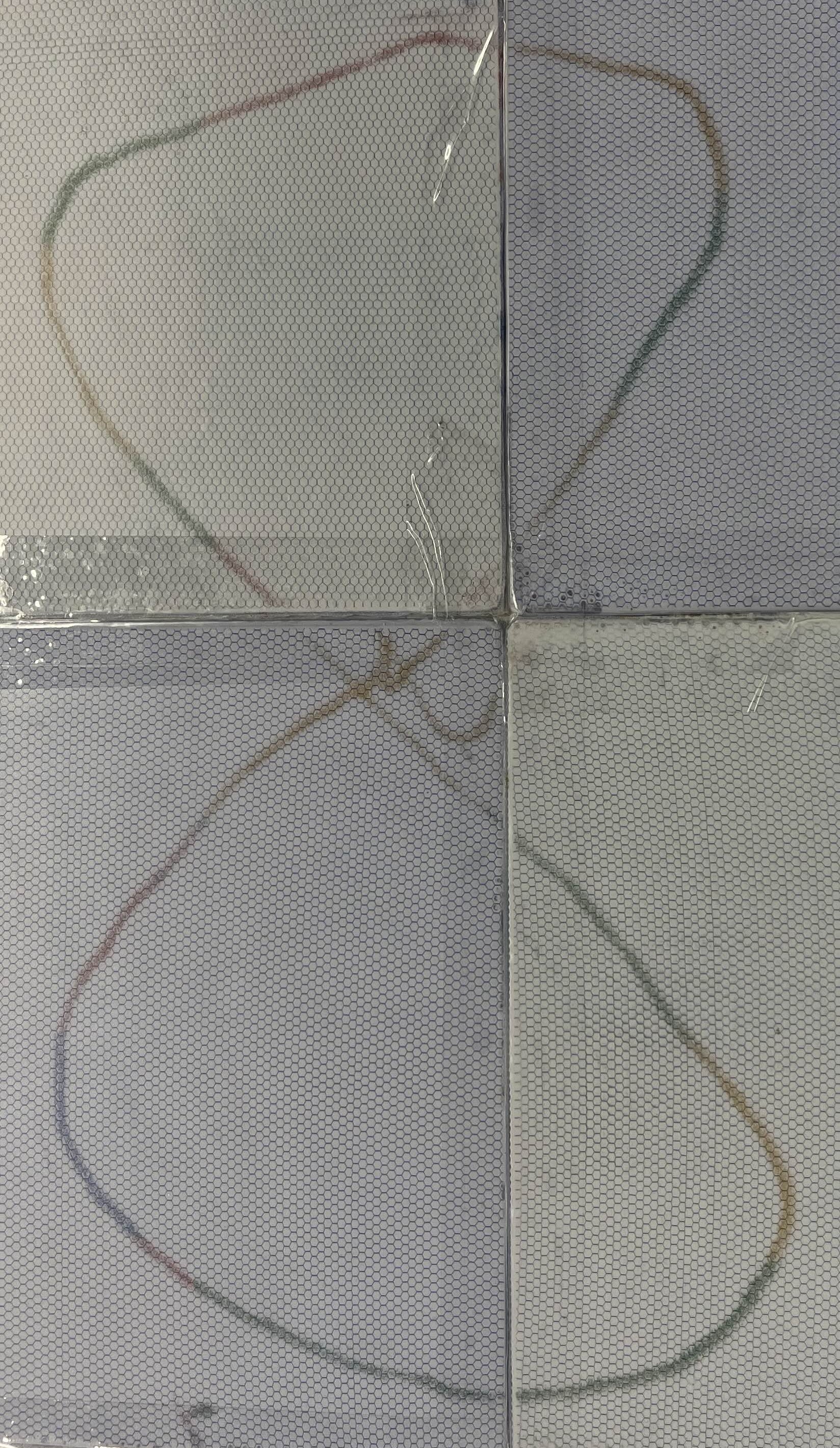}
        \caption{Magnet Dynamics}
        \label{fig:fiasda_magnet}
    \end{subfigure}
    \caption{Actual drawings on the board achieved by the HL Commander, LL Commander without magnet dynamics, and LL Commander with magnet dynamics controllers. This clearly depicts the significant smoothing provided by our LL Commander controller.}
    \label{fig:combined_fig8_on_board}
\end{figure}

As shown in table \ref{tab:fig8_errors}, HL commander produced less error than our controller. This is because HL commander moved along discontinuous coordinates at a fixed velocity, which enabled it to achieve closer accuracy at sharp turns. However, the MPC generated trajectory produced smoother drawings since velocity and attitude commands were passed to low level commander. This resulted in more appealing drawings despite the lower accuracy, observed in Figures \ref{fig:combined_fig8} and \ref{fig:combined_fig8_on_board}. The inclusion of magnet dynamics improved the results significantly, both in terms of x-, y-, z- error and aesthetic appeal of the drawings. This commander thus provides the additional benefit of full state control with comparable error and improved visual results.

\subsection{Complex Shapes}
\begin{figure}[H]
    \centering
    \begin{subfigure}[b]{0.5\linewidth}
        \centering
        \includegraphics[width=\linewidth]{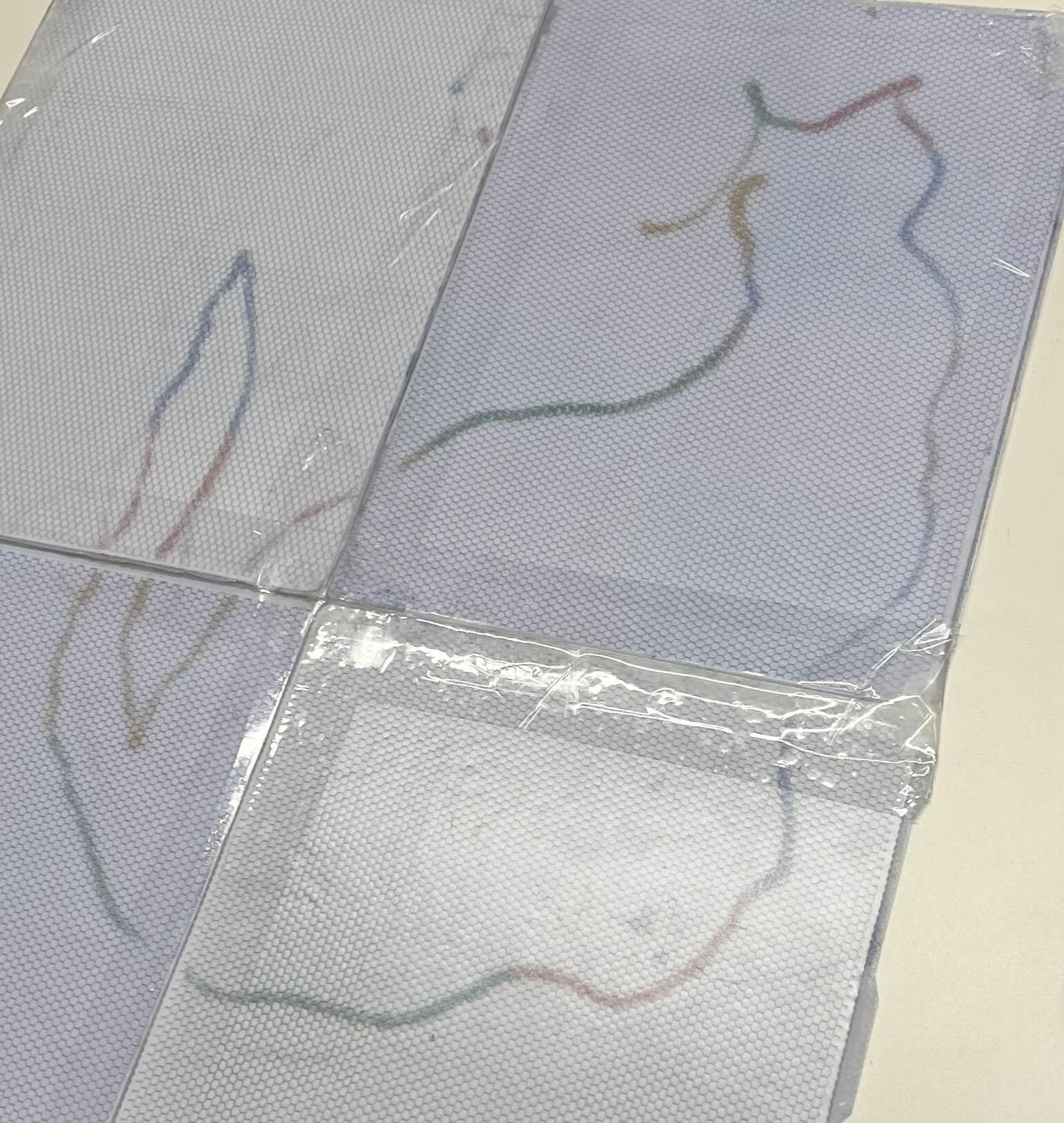}
        \caption{Cat}
        \label{fig:cat}
    \end{subfigure}
    \hfill
    \begin{subfigure}[b]{0.45\linewidth}
        \centering
        \includegraphics[width=\linewidth]{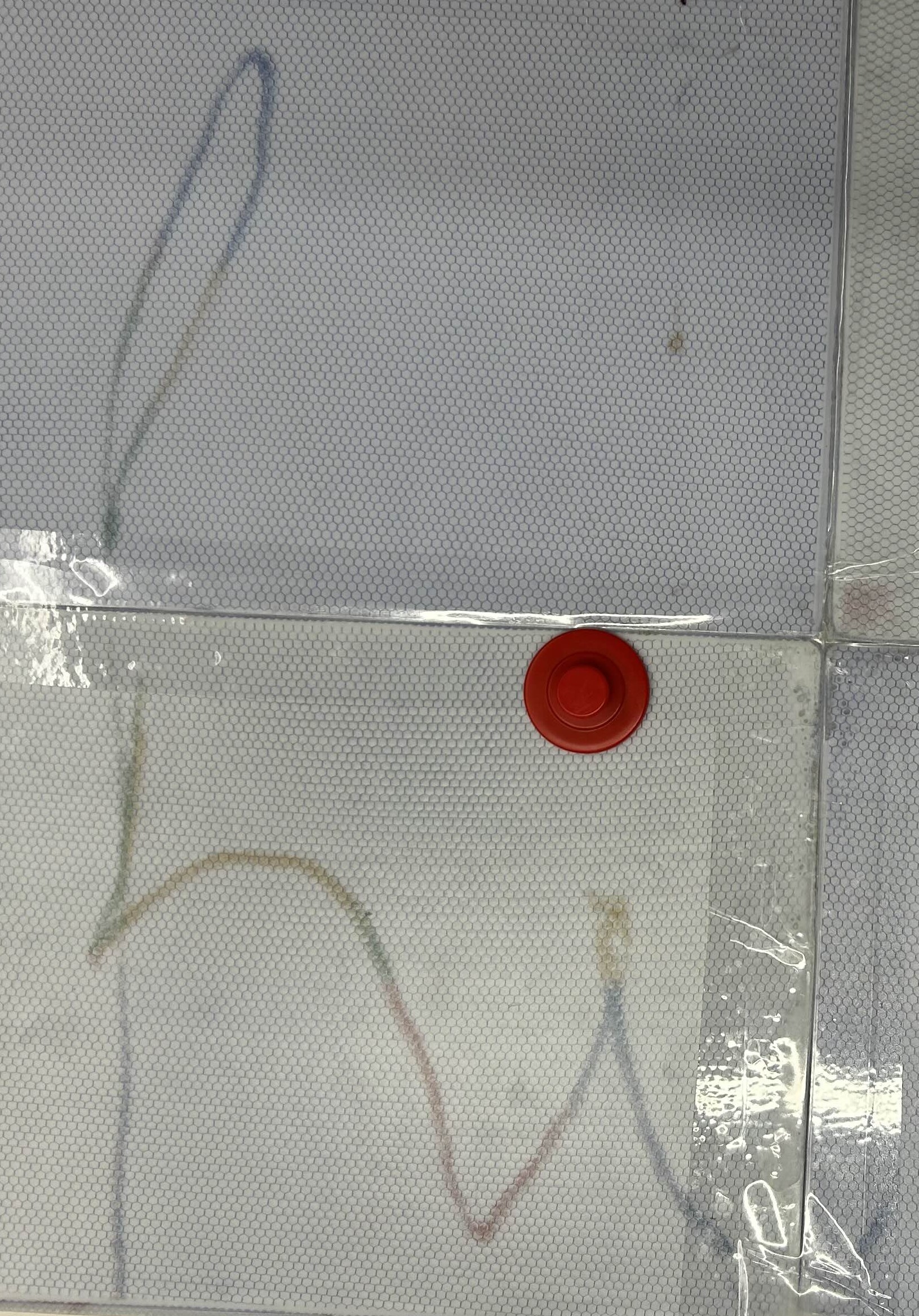}
        \caption{"hi"}
        \label{fig:hi}
    \end{subfigure}
 \caption{MPC with magnet dynamics was able to produce unique shapes.}
\end{figure}

Initially, tracking more complex shapes was found to result in larger error and smoothing of sharp edges, thought to be a limitation of the MPC formulation. This was initially tackled by increasing dt (65 Hz), the time to reach each waypoint. This proved effective in simulation. However, in hardware, this resulted in large swinging motions by LL Commander as it was unable to stabilize at each setpoint for longer time. However, increasing the number of waypoints (from 1000 to 1400) instead allowed for tracking these shapes, resulting in more detailed images, such as those in Figure~\ref{fig:cat} and Figure~\ref{fig:hi}.

\section{Conclusions}
Magnasketch was able to deliver a system capable of converting user input drawings to an optimal trajectory for a Crazyflie 2.0. Drawing was able to be demonstrated via a designed magnetic apparatus and magnetic board. MPC formulation with magnet dynamics paired with full state LL Commander control proved to have comparable performance to x, y, z HL Commander during experimental testing. 

Analyzing figure-8, circle, and cloud trajectories resulted in average errors of 3.9 cm, 4.4 cm, and 0.5 cm in x, y, and z respectively. Although the actual quantified error in HL Commander was less, when observing visualizations and the drawings themselves, our implementation was able to provide smoother, more aesthetically pleasing drawings due to following optimal, dynamically feasible paths, while also providing the benefit of full state control.

Error in executing complex paths was tackled by increasing the number of waypoints of input trajectories. This allowed our controller to be able to track differentiable and non-differentiable paths.

Other sources of error were also considered, which can be tackled in future work. First, MPC relies heavily on model dynamics. The magnet dynamics were a simplified estimate of the actual magnetic force. More intensive system identification, including of mass components, deadzones, and constraints of the system, could further improve MPC performance. 

LL Commander appeared to have limits unrelated to board contact. Slight error was still observed during trajectories followed above the board. This could have arisen to Lighthouse noise or the 100 Hz frequency limit on the Bitcraze Cascaded PID controller. The magnetic apparatus itself could be further iterated to handle Z-error. There also appeared to be a brief restabilization period when switching between LL and HL Commanders.

Addressing these in future work could further improve the performance of the Magnasketch drone.

\section{Appendix}
Code and sample shape drawings can be found \href{https://github.com/scott-wade/crazyflie_sketcher}{here}

\begin{figure*}[hbt!]
    \centering
    \includegraphics[width=0.9\textwidth]{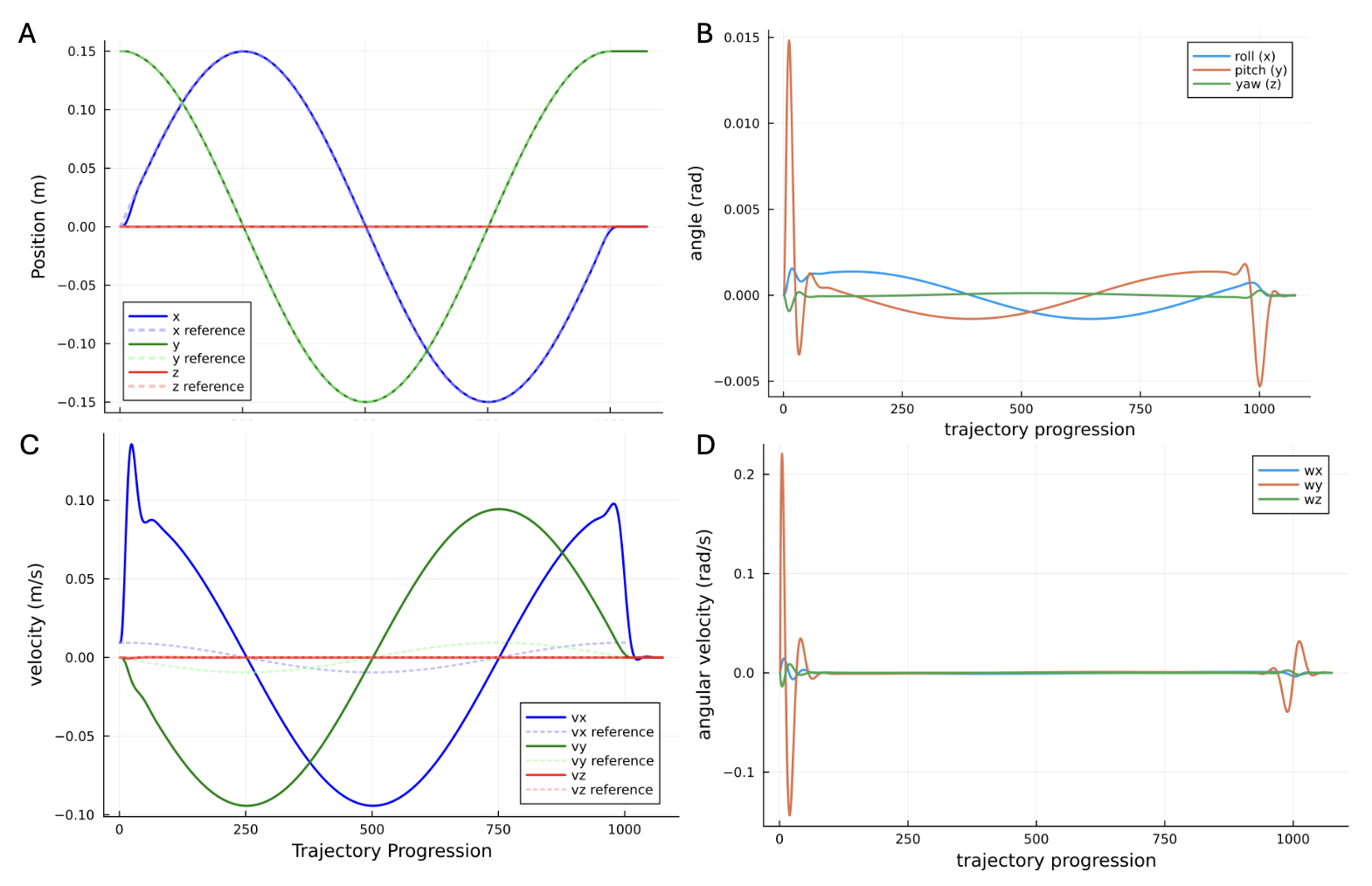}
    \caption{ MPC was used to generate an optimal, dynamically feasible trajectory tracked from a circle drawing. A simulation time step of 0.01 was set, 1000 points were in the drawing trajectory, and an MPC horizon of 75 steps was used. Magnet dynamics considerations were integrated into the nonlinear dynamics. A) Position states track the reference drawing accurately. The MPC was asked to track 0 z-height, as this was set as an offset in low level commander onboard the Crazyflie. B) Orientation states are shown in euler angles and demonstrate that only small changes in orientation occur to track the drawing in xy. This supports the use of a single hover linearization. C) The velocity states do not track the input drawing as tightly, however the drawing was meant only as an estimate, or warm start. MPC provided dynamically feasible velocity transitions. D) Angular velocity variations correlated with changes in orientation.}
    \label{circle_magnet}
\end{figure*}

\begin{figure*}[hbt!]
    \centering
    \includegraphics[width=0.9\textwidth]{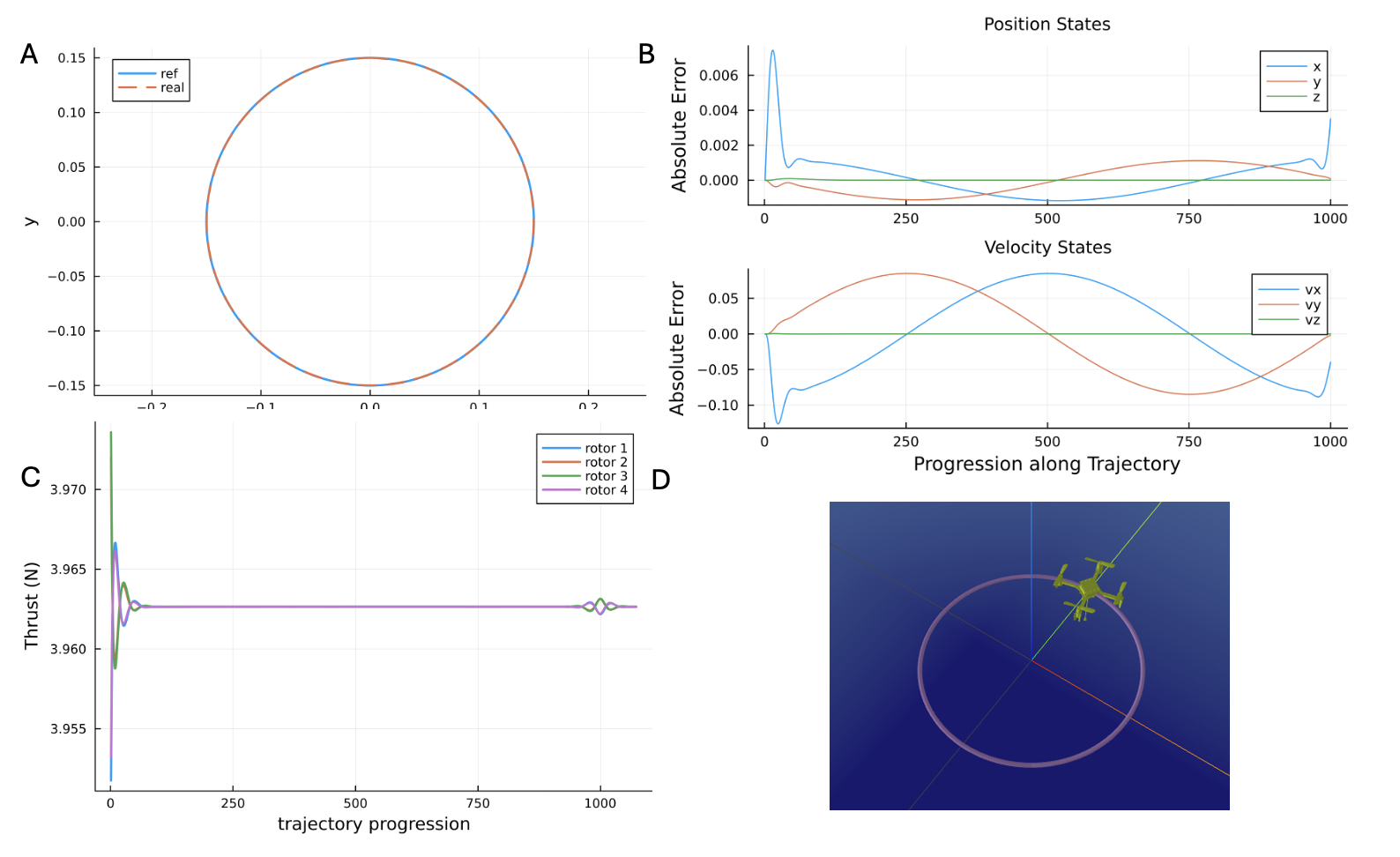}
    \caption{MPC was used to generate an optimal, dynamically feasible trajectory tracked from a circle drawing. A simulation time step of 0.01 was set, 1000 points were in the drawing trajectory, and an MPC horizon of 75 steps was used. Magnet dynamics considerations were integrated into the nonlinear dynamics. A) An xy state history plot shows accurate position tracking by the MPC trajectory. Because the curve is fully differentiable, there are no real devaitions from the given drawing. B) Position and velocity were input as nonzero by the drawing trajectory. The absolute error between the input position and the MPC-generated position states stays below 3mm except at the start of the trajectory when tracking begins. The velocity error between the drawing and the MPC trajectory are higher. Deviations from position and velocity were set in Q as 0.01m and 0.5m/s, respectively. C) Control effort stays around the required thrust to maintain hover. D) A screen shot of the mesh cat simulator shows the drone object tracking the circle drawing.}
    \label{circle_tracking_image}
\end{figure*}

\begin{figure*}[hbt!]
    \centering
    \includegraphics[width=0.9\textwidth]{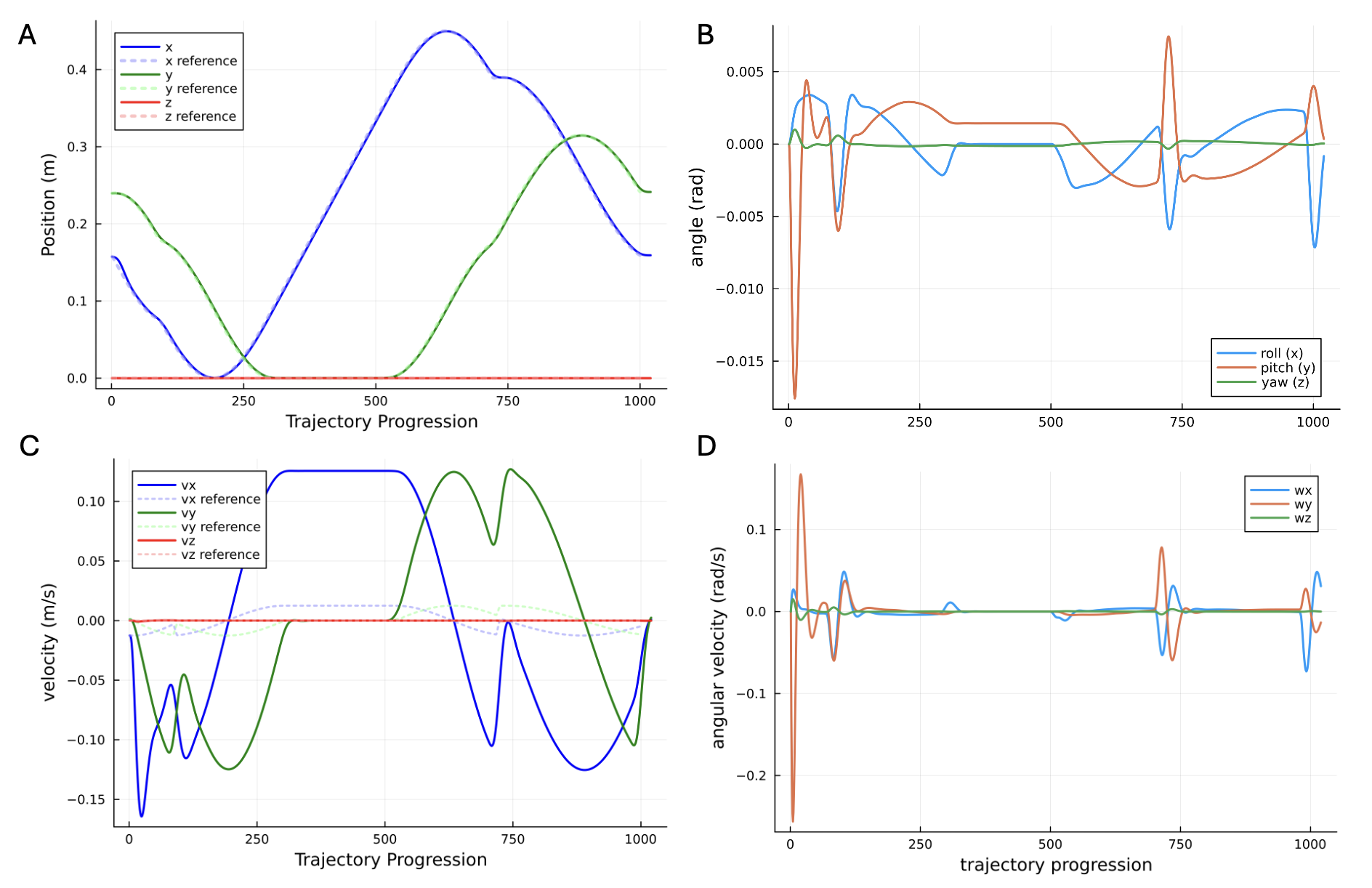}
    \caption{MPC was used to generate an optimal, dynamically feasible trajectory tracked from a cloud drawing. A simulation time step of 0.01 was set, 1000 points were in the drawing trajectory, and an MPC horizon of 20 steps was used. Magnet dynamics considerations were integrated into the nonlinear dynamics. A) Position states track the reference drawing accurately. The MPC was asked to track 0 z-height, as this was set as an offset in low level commander onboard the Crazyflie. B) Orientation states are shown in euler angles and demonstrate that only small changes in orientation occur to track the drawing in xy. This supports the use of a single hover linearization. C) The velocity states do not track the input drawing as tightly, however the drawing was meant only as an estimate, or warm start. MPC provided dynamically feasible velocity transitions. D) Angular velocity variations correlated with changes in orientation. Spikes correlate with sudden direction changes in the non-differentiable parts of the drawing.}
    \label{cloud_magnet}
\end{figure*}

\begin{figure*}
    \centering
    \includegraphics[width=0.9\textwidth]{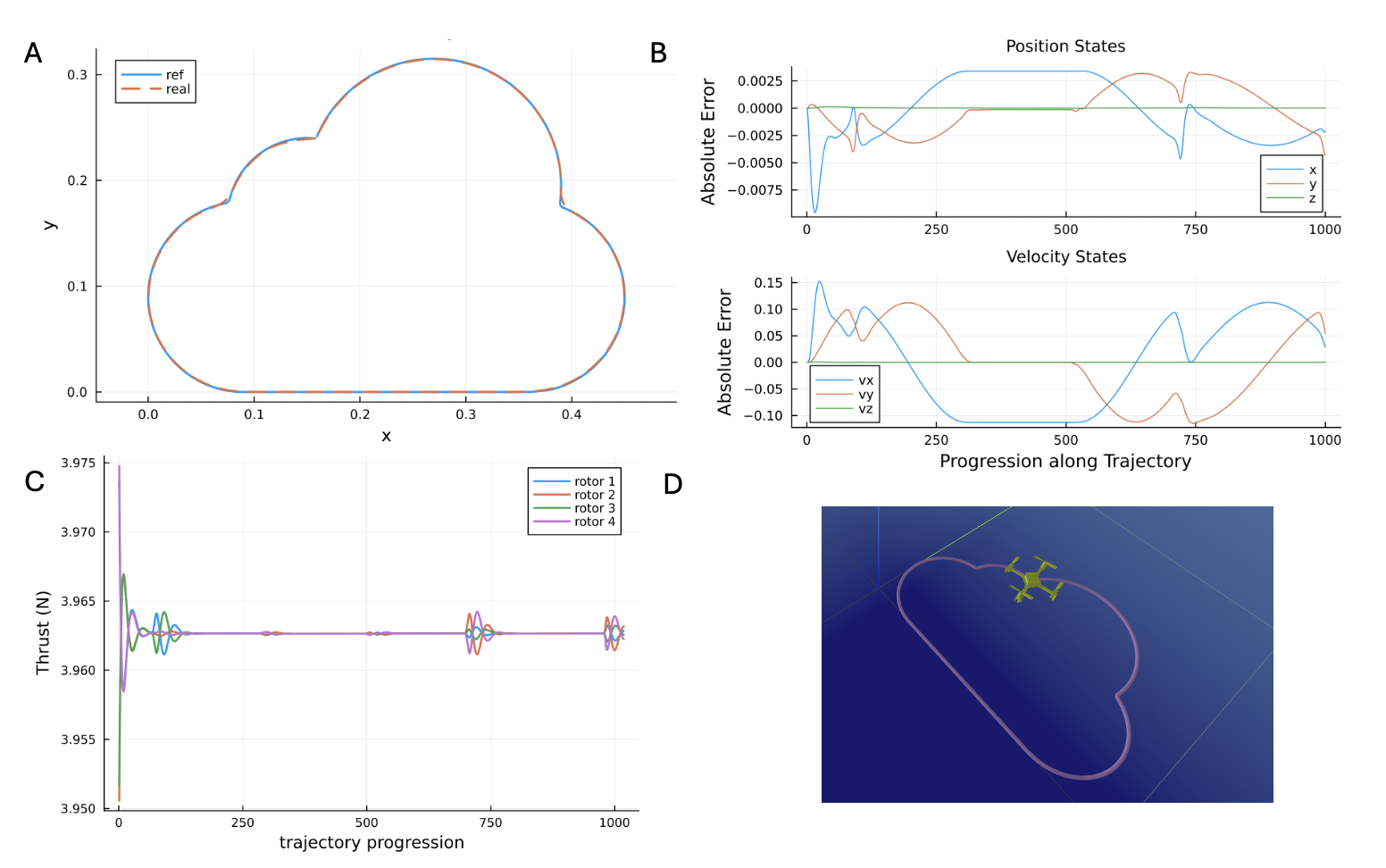}
    \caption{MPC was used to generate an optimal, dynamically feasible trajectory tracked from a cloud drawing. A simulation time step of 0.01 was set, 100 points were in the drawing trajectory, and an MPC horizon of 20 steps was used. Magnet dynamics considerations were integrated into the nonlinear dynamics. A) An xy state history plot shows accurate position tracking by the MPC trajectory. There are slight mismatches/curving happening at the non-differentiable portions of the drawing. B) Position and velocity were input as nonzero by the drawing trajectory. The absolute error between the input position and the MPC-generated position states stays below 3mm except at the start of the trajectory when tracking begins. The velocity error between the drawing and the MPC trajectory are higher. Deviations from position and velocity were set in Q as 0.01m and 0.5m/s, respectively. C) Control effort stays around the required thrust to maintain hover. D) A screen shot of the mesh cat simulator shows the drone object tracking the cloud drawing.}
    \label{cloud_tracking_image}
\end{figure*}

\begin{figure*}[hbt!]
    \centering
    \includegraphics[width=0.9\textwidth]{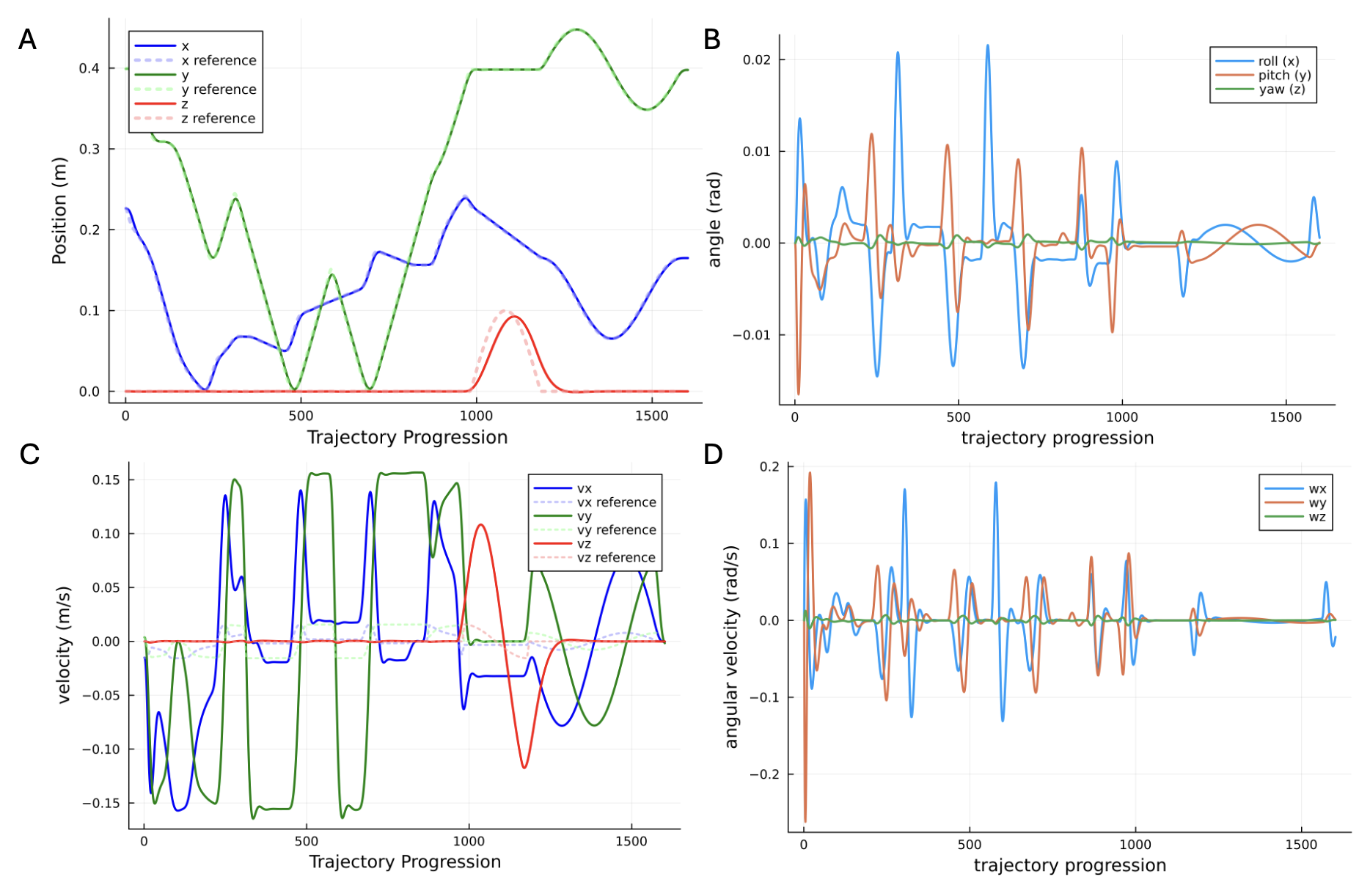}
    \caption{MPC was used to generate an optimal, dynamically feasible trajectory tracked from a drawing of a human. A simulation time step of 0.01 was set, 1582 points were in the drawing trajectory, and an MPC horizon of 20 steps was used. Magnet dynamics considerations were integrated into the nonlinear dynamics. A) Position states track the reference drawing accurately. The MPC was asked to track 0 z-height, as this was set as an offset in low level commander onboard the Crazyflie. B) Orientation states are shown in euler angles and demonstrate that only small changes in orientation occur to track the drawing in xy even when the . This supports the use of a single hover linearization. C) The velocity states do not track the input drawing as tightly, however the drawing was meant only as an estimate, or warm start. MPC provided dynamically feasible velocity transitions. D) Angular velocity variations correlated with changes in orientation. Spikes correlate with sudden direction changes in the non-differentiable text curve. Even with a challening drawing such as this, if enough points are given, the MPC trajectory provides a full state trajectory that accurately tracks the image.}
    \label{human_magnet}
\end{figure*}

\begin{figure*}[hbt!]
    \centering
    \includegraphics[width=0.9\textwidth]{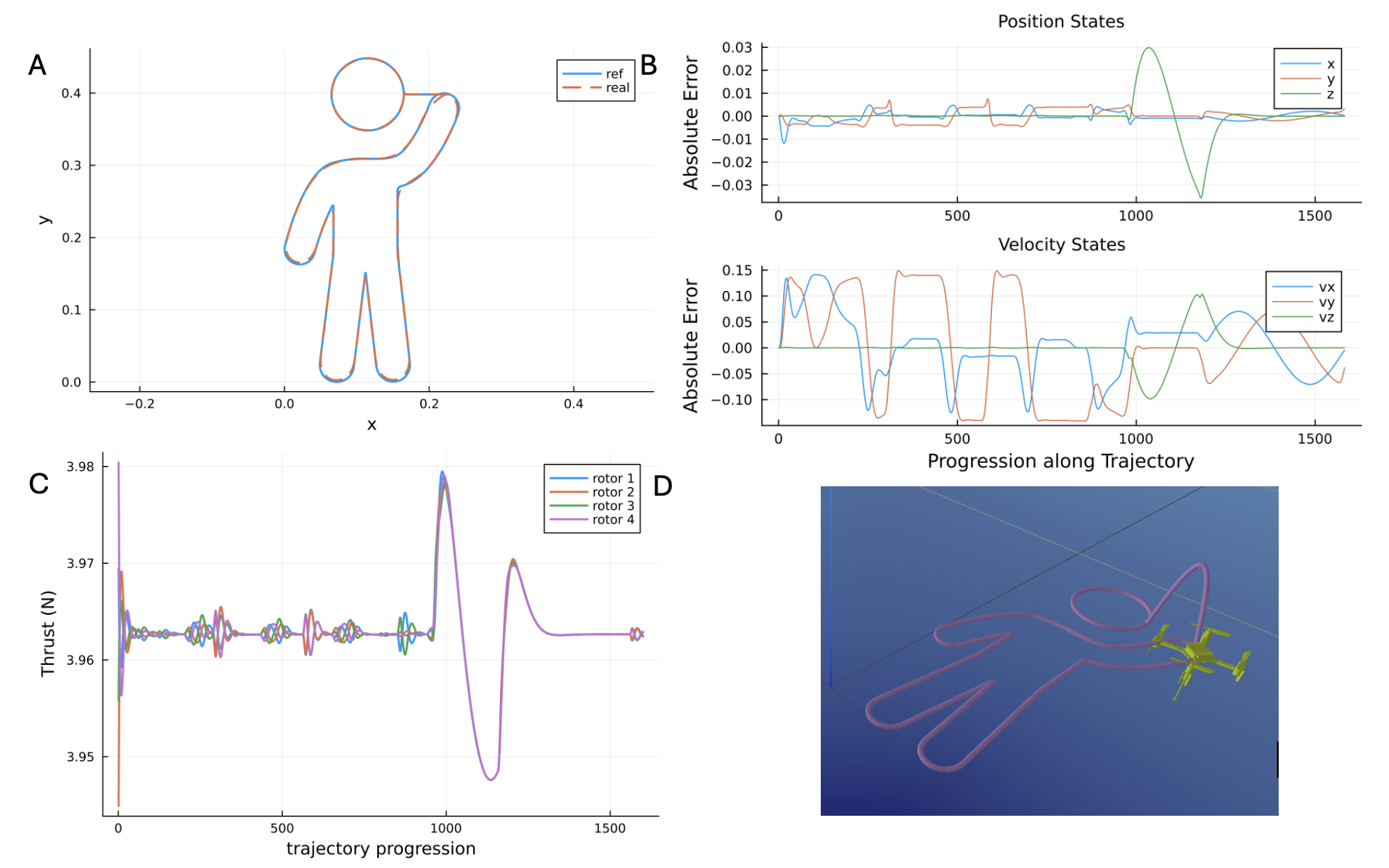}
    \caption{MPC was used to generate an optimal, dynamically feasible trajectory tracked from a drawing of a human. A simulation time step of 0.01 was set, 1582 points were in the drawing trajectory, and an MPC horizon of 20 steps was used. Magnet dynamics considerations were integrated into the nonlinear dynamics. A) An xy state history plot shows accurate position tracking by the MPC trajectory. There are slight mismatches/curving happening at the portions of the drawing where there is a tight corner or a sudden change. It is also important to note that the line between the hand and the head is actually completed off of the board and requires the drone to lift up. B) Position and velocity were input as nonzero by the drawing trajectory. The absolute error between the input position and the MPC-generated position states stays below 3mm except at the start of the trajectory when tracking begins. The velocity error between the drawing and the MPC trajectory are higher. Deviations from position and velocity were set in Q as 0.01m and 0.5m/s, respectively. C) Control effort stays around the required thrust to maintain hover except for when the drone is asked to lift off of the board and move to another location to draw the head. D) A screen shot of the mesh cat simulator shows the drone object tracking the drawing of the human.}
    \label{human_tracking_image}
\end{figure*}

\begin{figure*}[hbt!]
    \centering
    \begin{subfigure}[b]{0.32\linewidth}
        \centering
        \includegraphics[width=\linewidth]{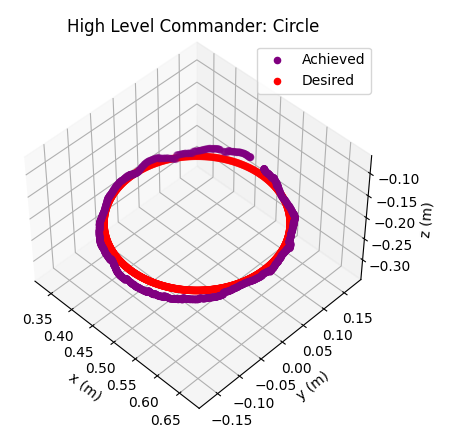}
        \caption{HL Commander}
        \label{fig:circle_HL}
    \end{subfigure}
    \hfill
    \begin{subfigure}[b]{0.32\linewidth}
        \centering
        \includegraphics[width=\linewidth]{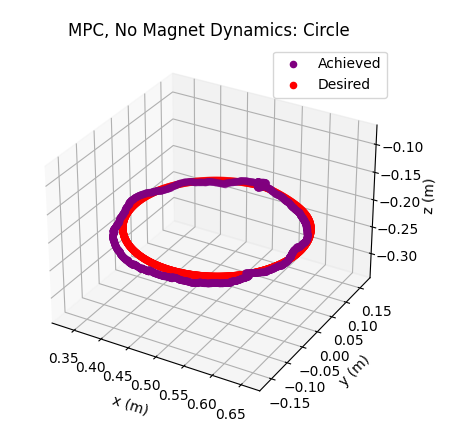}
        \caption{No Dynamics}
        \label{fig:circle_no_mag}
    \end{subfigure}
    \hfill
    \begin{subfigure}[b]{0.32\linewidth}
        \centering
        \includegraphics[width=\linewidth]{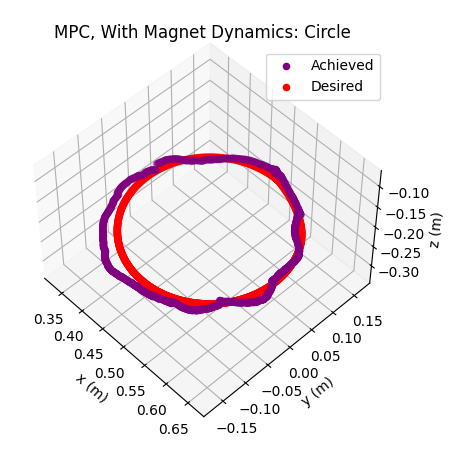}
        \caption{Magnet Dynamics}
        \label{fig:circle_magnet}
    \end{subfigure}
    \caption{Circle trajectory following by the HL Commander, LL Commander without magnet dynamics, and LL Commander with magnet dynamics controllers.}
    \label{fig:combined_circle}
\end{figure*}

\begin{figure*}[hbt!]
    \centering
    \begin{subfigure}[b]{0.32\linewidth}
        \centering
        \includegraphics[width=\linewidth]{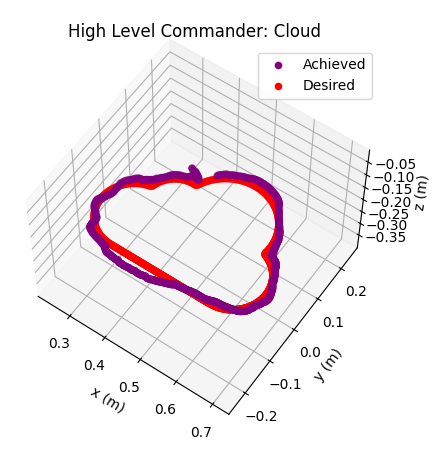}
        \caption{HL Commander}
        \label{fig:cloud_HL}
    \end{subfigure}
    \hfill
    \begin{subfigure}[b]{0.32\linewidth}
        \centering
        \includegraphics[width=\linewidth]{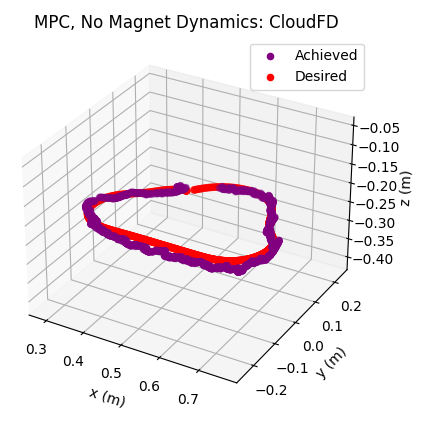}
        \caption{No Dynamics}
        \label{fig:cloud_no_mag}
    \end{subfigure}
    \hfill
    \begin{subfigure}[b]{0.32\linewidth}
        \centering
        \includegraphics[width=\linewidth]{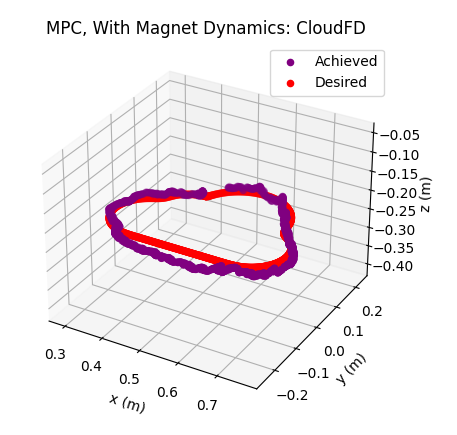}
        \caption{Magnet Dynamics}
        \label{fig:cloud_magnet}
    \end{subfigure}
    \caption{Cloud trajectory following by the HL Commander, LL Commander without magnet dynamics, and LL Commander with magnet dynamics controllers.}
    \label{fig:combined_cloud}
\end{figure*}
\begin{figure*}[hbt!]
    \centering
    \begin{subfigure}[b]{0.45\linewidth}
        \centering
        \includegraphics[width=\linewidth]{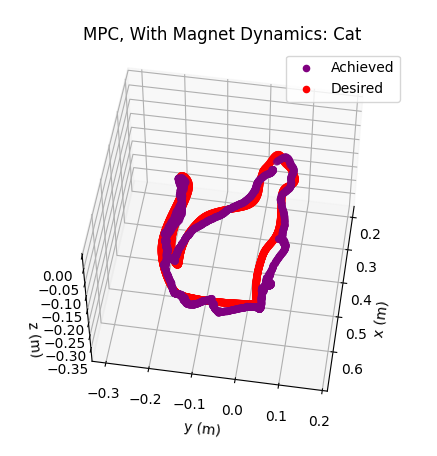}
        \caption{Cat}
        \label{fig:cat_plot}
    \end{subfigure}
    \hfill
    \begin{subfigure}[b]{0.45\linewidth}
        \centering
        \includegraphics[width=\linewidth]{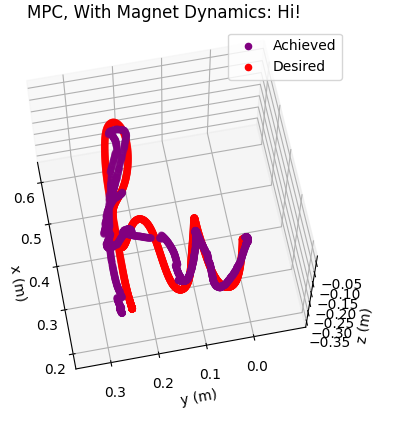}
        \caption{"Hi"}
        \label{fig:hi_plot}
    \end{subfigure}
    \caption{Additional trajectory following by the HL Commander, LL Commander without magnet dynamics, and LL Commander with magnet dynamics controllers, demonstrating the ability to follow non-differentiable trajectories.}
\end{figure*}



\newpage
\bibliography{references}
\vspace{12pt}
\bibliographystyle{IEEEtran}

\end{document}